\PassOptionsToPackage{colorlinks}{hyperref}
\documentclass[sigconf,letter,10pt,screen]{acmart}
\definecolor[named]{ACMPurple}{cmyk}{0,0.76,0.29,0.24} 
\definecolor[named]{ACMDarkBlue}{cmyk}{0,0.76,0.29,0.24}

\usepackage[english]{babel}
\usepackage{xspace}
\usepackage{xcolor}
\usepackage{soul}
\usepackage{multirow}
\usepackage{tabularx}
\usepackage{graphicx}
\usepackage{subfig}
\usepackage{siunitx}
\usepackage{wasysym}
\usepackage[most]{tcolorbox}
\usepackage{flushend}

\usepackage{courier}
\usepackage{listings}
\makeatletter
\newcommand{\srcsize}{\@setfontsize{\srcsize}{5.1pt}{5.1pt}}
\makeatother
\usepackage{xcolor}

\definecolor{codegreen}{rgb}{0,0.6,0}
\definecolor{codegray}{rgb}{0.5,0.5,0.5}
\definecolor{codepurple}{rgb}{0.58,0,0.82}
\definecolor{backcolour}{rgb}{0.95,0.95,0.92}

\renewcommand\footnotetextcopyrightpermission[1]{} 

\newif\ifcomment
\commentfalse

\ifcomment

\newcounter{AFNumberOfComments}
\stepcounter{AFNumberOfComments}
\newcommand{\afnote}[1]{\textcolor{red}{\small AF-\arabic{AFNumberOfComments}\stepcounter{AFNumberOfComments}: #1}}

\newcounter{JKNumberOfComments}
\stepcounter{JKNumberOfComments}
\newcommand{\jknote}[1]{\textcolor{olive}{\small JK-\arabic{JKNumberOfComments}\stepcounter{JKNumberOfComments}: #1}}

\newcounter{JMNNumberOfComments}
\stepcounter{JMNNumberOfComments}
\newcommand{\jmnnote}[1]{\textcolor{blue}{\small JMN-\arabic{JMNNumberOfComments}\stepcounter{JMNNumberOfComments}: #1}}

\newcounter{CWNumberOfComments}
\stepcounter{CWNumberOfComments}
\newcommand{\cwnote}[1]{\textcolor{magenta}{\small CW-\arabic{CWNumberOfComments}\stepcounter{CWNumberOfComments}: #1}}

\else
\newcommand\afnote[1]{}
\newcommand\jknote[1]{}
\newcommand\jmnnote[1]{}
\newcommand\cwnote[1]{}
\fi
\usepackage{fancyhdr}
\fancypagestyle{plain}{%
\fancyhf{} %
\fancyfoot{}%
}
\hypersetup{
colorlinks = true,
citecolor = {magenta},
linkcolor = {black},
}

\newcommand{\ISCX}{$\mathtt{ISCX}$\xspace}
\newcommand{\ISCXVPN}{$\mathtt{ISCX\text{-}VPN}$\xspace}
\newcommand{\ISCXTOR}{$\mathtt{ISCX\text{-}Tor}$\xspace}
\newcommand{\UCDAVIS}{$\mathtt{UCDAVIS}19$\xspace}

\newcommand{\MIRAGEA}{$\mathtt{MIRAGE}19$\xspace}
\newcommand{\MIRAGEB}{$\mathtt{MIRAGE}22$\xspace}
\newcommand{\UTMOBILENET}{$\mathtt{UTMOBILENET}21$\xspace}
\newcommand{\APPCLASSNET}{$\mathtt{AppClassNet}\xspace$}
\newcommand{\CESNETTLS}{$\mathtt{CESNET}$-$\mathtt{TLS}$\xspace}
\newcommand{\SCRIPT}{$\mathtt{script}$\xspace}
\newcommand{\HUMAN}{$\mathtt{human}$\xspace}
\newcommand{\LEFTOVER}{$\mathtt{leftover}$\xspace}
\newcommand{\reddagger}{$^{\textcolor{red}\dagger}$}
\newcommand{\redddagger}{$^{\textcolor{red}\ddagger}$}
\newcommand{\redasterisc}{$\textcolor{red}{^*}$}

\newcommand{\TARGETPAPER}{$\fontsize{9}{10}\selectfont\mathsf{\textsc{Ref-Paper}}$\xspace}
\newcommand{\TARGETAUTHORS}{\emph{Horowicz}~et~al.\xspace}

\usepackage{tikz}

\newcommand{\tinytiny}[1]{\fontsize{6}{10}\selectfont \textcolor{gray}{$\pm #1$}}

\newtcolorbox{quotethepaper}{
colback=gray!5!white,
colframe=gray!75!black,
fonttitle=\bfseries,
enhanced,
breakable,
parbox=false,
skin first=enhanced,
skin middle=enhanced,
skin last=enhanced,
parbox=false,
size=small,
}

\newcommand{\mypar}[1]{\paragraph*{\textbf{#1}}}

\makeatletter
\def\@copyrightspace{\relax}
\makeatother

\begin{document}
\title{Replication: Contrastive Learning and Data Augmentation in Traffic Classification Using\\a Flowpic Input Representation}

\author{Alessandro Finamore}
\affiliation{
{\fontsize{10}{10}\selectfont Huawei Technologies SASU}
\country{France}
}
\email{{alessandro.finamore@huawei.com}}

\author{Chao Wang}
\affiliation{
{\fontsize{10}{10}\selectfont Huawei Technologies SASU}
\country{France}
}
\email{wang.chao3@huawei.com}

\author{Jonatan Krolikowski}
\affiliation{
{\fontsize{10}{10}\selectfont Huawei Technologies SASU}
\country{France}
}
\email{jonatan.krolikowski@huawei.com}

\author{Jose M. Navarro}
\affiliation{
{\fontsize{10}{10}\selectfont Huawei Technologies SASU}
\country{France}
}
\email{jose.manuel.navarro@huawei.com}

\author{Fuxing Chen}
\affiliation{
{\fontsize{10}{10}\selectfont Huawei Technologies SASU}
\country{France}
}
\email{chenfuxing@huawei.com}

\author{Dario Rossi}
\affiliation{
{\fontsize{10}{10}\selectfont Huawei Technologies SASU}
\country{France}
}
\email{dario.rossi@huawei.com}

\begin{abstract}
Over the last years we witnessed a renewed interest toward
Traffic Classification (TC) captivated by the rise of Deep
Learning (DL). Yet, the vast majority of TC literature lacks code artifacts, performance assessments across
datasets and reference comparisons against Machine
Learning (ML) methods. Among those works,
a recent study from IMC'22~\cite{horowicz2022imc-fewshotcl} is worth of attention
since it adopts recent DL methodologies (namely, \textit{few-shot} learning, \textit{self-supervision} via \textit{contrastive learning}  and \textit{data augmentation}) appealing for networking as they enable to learn from a few samples and transfer across datasets. The main result of \cite{horowicz2022imc-fewshotcl}  on the  \UCDAVIS, \ISCXVPN and \ISCXTOR datasets is that, with such DL methodologies,  100 input samples are enough to achieve very high accuracy using an input representation called ``flowpic'' (i.e.,  a per-flow 2d histograms of the packets size evolution over time).

In this paper (i) we \textit{reproduce} \cite{horowicz2022imc-fewshotcl} on the same datasets
and (ii) we \textit{replicate} its most salient aspect (the importance of data augmentation) on three additional public datasets (\MIRAGEA, \MIRAGEB and \UTMOBILENET).
While we confirm most of the original results, we also found a $\approx$20\% accuracy drop on some of the investigated scenarios
due to a data shift in the original dataset that we uncovered. Additionally, our study validates that the data augmentation strategies studied in \cite{horowicz2022imc-fewshotcl} perform well on other datasets too.
In the spirit of reproducibility and replicability we make all artifacts (code and data) available to the research community at \url{https://tcbenchstack.github.io/tcbench/}.
\end{abstract}

\begingroup
\mathchardef\UrlBreakPenalty=10000
\maketitle
\endgroup

\section{Introduction
\label{sec:introduction}
}

Traffic classification (TC) is a long investigated topic in the networking community with seminal works dating back nearly two decades ago~\cite{moore2005sigmetrics} which have been instrumental for bringing Machine Learning (ML) tools into networks operation and management.
Since then, the TC field has been flourishing with literature and it is regularly surveyed~\cite{armitage2008survey,pacheco2018survey}.
The recent hype of Deep Learning (DL) has expanded the interest on the field with several contributions from flagship ACM and IEEE conferences, including IMC~\cite{horowicz2022imc-fewshotcl}.

Despite the progress made, reproducing (and replicating) research can still be a challenge~\cite{reproducibility-survey}, especially for TC.
This is often rooted back in the well known self-awareness that ``a scientific publication is not the scholarship itself, it is merely advertising of the scholarship''~\cite{donoho1995wavelab}.
For the networking field, reproducibility has become more of a commonplace in the last decade, thanks to the emergence of tools (such as specialized containers~\cite{handigol2012conext-reproducible}), community-wide awareness (such as dedicated workshops~\cite{reproducibility-workshop17}) and policies (such as ACM badging~\cite{acm-badging}). Considering TC, difficulties are knowingly aggravated by data availability (yet, see Sec.\ref{sec:bg:labelscarcity} for a positive outlook) and pertinence (i.e., due to datasets bias~\cite{jacobs2022ccs} and ageing, which mandates replication across several datasets).

\begin{figure*}
\includegraphics[width=\textwidth]{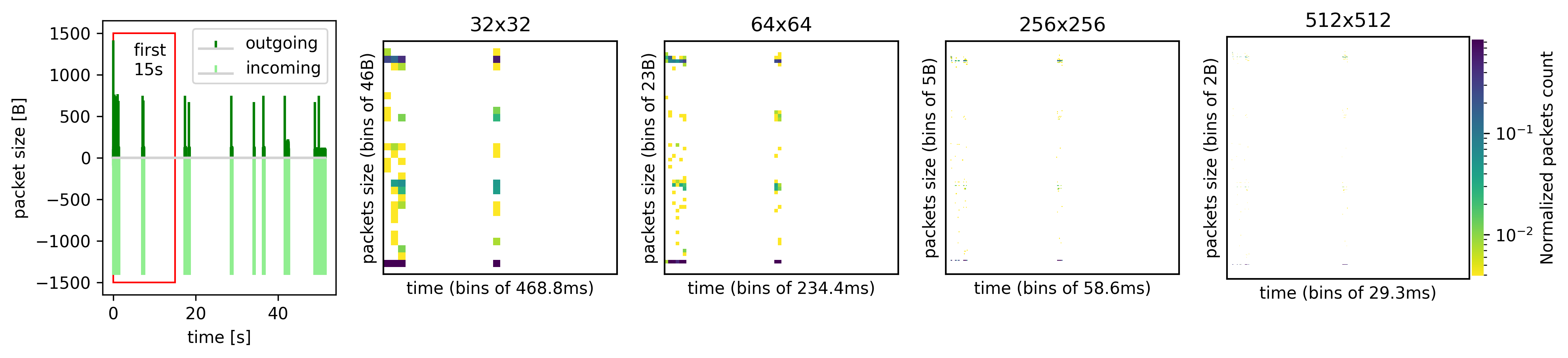}
\caption{
\label{fig:flowpic-example} Example of a packet time series transformed into a flowpic representation for a randomly selected YouTube flow in the \UCDAVIS dataset. Heatmaps are in a log scale
normalized between the max and min value for each flowpic, 
with higher packets count values having darker shades (images better viewed digitally).}
\end{figure*}

Following a different direction than previous literature, this paper aims to \textit{reproduce} and \textit{replicate} research results from a recent TC study. In other words, our final goal is not only to investigate the scientific aspects behind the methodologies under study per-se, but to enable the research community to take advantage
of our code base and artifacts such as models, logs, and curated datasets with a complementary website
to document and navigate the artifacts (see App.~\ref{app:artifacts}).

More in detail, we aim to reproduce the most important aspects of an interesting recent work on TC that appeared as a \emph{short paper} in IMC'22 program ~\cite{horowicz2022imc-fewshotcl}. This  study uses recent promising DL techniques (notably, \textit{few-shot} learning, \textit{self-supervision} via \textit{contrastive learning} and \textit{data augmentation}) that we believe to be worthy of community-wide interest as they relate to practical problems that plague ML and DL application for TC (e.g., poor model generalization and label scarcity).
Summarizing our main findings:
\begin{itemize}
\item Regarding \emph{reproducibility}, from a \emph{qualitative} viewpoint, we were able to reproduce most of the results of~\cite{horowicz2022imc-fewshotcl}, e.g., confirming the interest in  few-shot learning, self-supervision via contrastive loss and data augmentation;
from a \emph{quantitative} viewpoint, we incurred in unexpected results with large discrepancies that we were able to drill down and explain. Furthermore, with respect to the original publication, we calculated confidence intervals on our results to achieve better statistical validity, which may require tempering some of the observations in~\cite{horowicz2022imc-fewshotcl}. 

\item Concerning \emph{replicability}, we find qualitative agreement and confirm that the data augmentation policies selected in \cite{horowicz2022imc-fewshotcl} behave consistently on other datasets.

\item Philosophically, this work could be read as a chapter of the famous Queneau's   book ``Exercises in style''\cite{queneau},
cast to tell a  traffic classification story.
In particular, some aspects of the target paper (that we discuss in detail later) are missing despite being key for
effectively reproducing the
study.\footnote{
We carried out our study based on what is
reported in \cite{horowicz2022imc-fewshotcl}. In fact, due to the double blind policy of the submission,
we reached out multiple times to the original authors only in the weeks related to the shepherding of this article
but we received only short and delayed responses.
It is worth mentioning that we found a git repository~\cite{unofficialgit}
related to \cite{horowicz2022imc-fewshotcl} but
it presents several important limitations (see App.~~\ref{quote: icdm-datashit-explain})
and cannot be used to replicate~\cite{horowicz2022imc-fewshotcl}.
} Hence, readers must be aware that our resulting exercise is one of the many styles to tell the same story~\cite{queneau}.
\end{itemize}

In the following, we first provide background about the target study (Sec. \ref{sec:relatedwork}). We then lay out our replicability and reproducibility goals, providing information about artifacts (Sec. \ref{sec:methodology}). We continue by discussing our experimental protocol and results (Sec. \ref{sec:evaluation}) before concluding with final remarks (Sec. \ref{sec:conclusions}). Details that we believe to be needed to make the paper self contained are deferred to App.~\ref{app:architectures}--\ref{appendix:tukey}.

\section{Background and motivation
\label{sec:relatedwork}
}

Due to its nature, the present study falls in the broad area of reproducibility that started becoming a popular subject in networking a decade ago~\cite{handigol2012conext-reproducible}. Given its breadth, it is out of the scope of this paper to review the whole reproducibility discipline. Conversely, we focus on the
selected target paper, starting with an overview of its scope and contributions which we expand with relevant related work.

\subsection{Target paper}
In this work we replicate~\cite{horowicz2022imc-fewshotcl}, which in the remainder of this paper we will also refer to as  \TARGETPAPER or \TARGETAUTHORS
In this \emph{short} study, \TARGETAUTHORS quantify the performance of classification tasks
when using only up to 100 training samples (i.e., \emph{few shot learning} scenarios) yet
increasing the training dataset with synthetic samples created with data augmentation functions.
Authors consider both a supervised and an unsupervised setting, the latter represented by
the popular \emph{contrastive learning} approach named SimCLR~\cite{chen2020simple-simclr},
which starts from pre-training a model in an unsupervised fashion
and later fine-tunes it to address a target task using a small number of labeled samples.
Rather than using packet time series,
\TARGETAUTHORS use a \emph{flowpic input representation}, i.e., a 2d summary
of a network flow dynamics.
Overall, we identify three contributions from the \TARGETPAPER:
($i$) a benchmark of the effect on the performance of Convolutional Neural Network (CNN) models across 7 \emph{data augmentations} (Sec.~\ref{sec:bg:labelscarcity}), tested on two datasets, \UCDAVIS and \ISCX (Sec.~\ref{sec:datasets}); ($ii$) an evaluation of SimCLR (Sec.~\ref{sec:bg:contrastive}) and its sensitivity to the number of samples used during fine-tuning; ($iii$) an ablation of the fine-tuning performance when using alternative input formats to flowpic  (Sec.~\ref{sec:bg:input}).
In the remainder of this section we expand on each of those contributions.

\subsection{Input data representation}\label{sec:bg:input}
\mypar{Target work}
The flowpic representation used in the 
$\fontsize{9}{10}\selectfont\mathsf{\textsc{Ref-}}$ $\fontsize{9} {10}\selectfont\mathsf{\textsc{Paper}}$\xspace
was originally introduced by (part of) the same authors at an INFOCOM'19 workshop~\cite{shapira2019infocom-flowpic}. In Figure~\ref{fig:flowpic-example} we
show a YouTube flow (extracted randomly from the \UCDAVIS dataset) as well as its related
flowpic at different resolutions. The left most plot shows the packet time series.
Notice the expected bursty nature typical of video streaming services. The \TARGETPAPER computes a flowpic using only the
first 15s of the time series. Specifically, both the 15s and the packets size range (0-1500) are split into bins
based on the resolution of the target flowpic.\footnote{Traffic directionality is not considered when composing the flowpic in the \TARGETPAPER although the representation could be reformulated to take it into account.}
For instance a 32$\times$32 flowpic leads to 469.8ms time bins and 46B packet size bins.
Then, the count of the packets occurring in each time window are tallied based on the defined packet size bins. In other words,
each time window provides a frequency histogram of the packet sizes, and by vertically stacking all the histograms
we obtain a ``picture'' of the flow dynamics. For instance, at the 32$\times$32 resolution,
the vertical stripes match the packet bursts of the original time series.
This sort of patterns make the flowpic representation appealing for CNN-based DL architectures
as convolutional layers are explicitly designed to extract features to detect such patterns.
Yet, the higher the flowpic resolution, the sparser the representation,
and the higher their computational process. While flowpic was introduced with a 1500$\times$1500
resolution, in the \TARGETPAPER this is compared against a 32$\times$32 resolution, i.e., a mini-flowpic.

\mypar{Related work}
Despite being well suited for CNN architectures, the flowpic representation
is not a mainstream choice for TC as it requires to observe multiple seconds of traffic.
This can enforce a late/post-mortem classification (i.e., after the flow ends), which, while still useful for monitoring, might not fit network management needs---prioritization, scheduling and shaping benefit from classification after the first few packets, so waiting for multiple seconds to take action can be sub-optimal. Conversely, the most common input features
in the traffic classification  literature are \textit{packet time series}
(e.g., the size, direction, inter-arrival time of the first
10 packets of a flow) and \textit{payload bytes}
(e.g., the first 784 L4 payload bytes).
Since time series (and payload) features enable early classification, they have been the go-to choice since seminal works~\cite{crotti2007ccr,bernaille2006ccr,moore2005pam}. Additionally, time series and payload input can be combined in ``multi modal'' architectures~\cite{aceto2019comnet-mimetic,orange2021sigmetrics,luxemburk2023comnet-cesnettls} that have become popular in networking and other fields---rather than selecting either one representation or the other, a DL model can be designed to learn from different input formats at the same time.
While we acknowledge it would be interesting to benchmark architectures and input representation, it is beyond the scope of our reproducibility/replicability study.

\subsection{Label scarcity and data augmentation}\label{sec:bg:labelscarcity}

\mypar{Target work}
Supervised learning requires large labeled datasets. As these are notoriously difficult to share and labeling is costly, the ability to learn from as few labeled samples as possible is particularly appealing.

In this direction, the \TARGETPAPER
considered two mid-sized datasets---\UCDAVIS~\cite{rezaei2019ICDM-ucdavis}, which
contains 5 classes with up to $\approx$1,000 samples per class, and
\ISCXVPN and \ISCXVPN~\cite{iscx-datasets}, which were processed and combined to
obtain 10 classes with a few flows each---and investigated the classification performance when
\emph{learning from only 100 samples}. To do so, \TARGETAUTHORS considered \emph{data augmentation} techniques
applied to either flowpics (e.g., rotation) or
to the packets time series (e.g., altering inter-arrival times)
from which the flowpics are then computed.
The \TARGETPAPER shows that simple data augmentations can indeed be beneficial even when using a few samples (with authors preferring time series transformations over image-related ones). These are interesting findings worth reproducing---on flowpic in the context of this work---and we believe they should be extended to packet time-series too in a future work.

\mypar{Related work}
Concerning \emph{label scarcity}, until recent times only small-sized datasets for TC were publicly available---tens of classes and a few thousands of flows at best~\cite{rezaei2019ICDM-ucdavis,iscx-datasets}. This situation changed with the release of significantly larger datasets---hundreds of classes and millions of flows~\cite{wang2022ccr-appclassnet,luxemburk2023comnet-cesnettls,luxemburk2023datainbrief-cesnet-quic}.
Whereas this encouraging trend is slowly making TC public datasets reach the scale
of computer vision datasets, reducing dependency from labels is still a desirable goal and an
open question for the ML research community as a whole.

Concerning \textit{ data augmentation} we find that, while it is widely adopted in computer vision (from the very first work at the root of the hype of CNNs in the field \cite{krizhevsky2012imagenet}),
only a handful of studies use it in TC literature.
Particularly relevant is \cite{rezaei2019ICDM-ucdavis}
where the authors, beside introducing the \UCDAVIS dataset,
use as input packet time series, which are sampled into ``subflows''. Specifically,
given the packet series (size, direction and inter arrival time),
the authors propose to sample values based on different
policies (e.g., selecting one packet every N from a random starting
point); hence, from one flow they obtain multiple ``subflows''
which semantically correspond to a coarser-grained ``view'' of the original
flow. The authors use \UCDAVIS and \ISCXVPN with a
two-step learning process: first, they \emph{pre-train}
a model in an unsupervised manner, targeting a 24-way regression problem from the generated subflows, i.e., they create a model that, given a subflow as input, can provide 24 metrics extracted from the flow;
then such model is \emph{fine-tuned} to obtain the target classifier using up to 20 labeled
samples per class. In other words, \cite{rezaei2019ICDM-ucdavis}
uses the two-step training approach of the \TARGETPAPER
but does not adopt contrastive learning.
Another recent study about data augmentation is
\cite{auwal-access20-dcgan} where the authors use
a GAN-based approach that learns to augment packet time series
during training. Conversely, in the \TARGETPAPER the augmentations
are designed based on domain knowledge.
Overall, while we believe that a broader and more systematic (i.e., across multiple datasets and inputs) comparison of data augmentation techniques in the TC field should be of community-wide interest, in our study we use the same point of view of the \TARGETPAPER.

\subsection{Contrastive learning}\label{sec:bg:contrastive}

\mypar{Basic principles}
Contrastive learning is particularly relevant and among the main reasons for our selection of the \TARGETPAPER.
However, to understand this we first need to review some basic principles of DL model training.

At a high level, a supervised DL classifier is typically a composition of a feature extractor and a linear classifier.
During training, the feature extractor learns how to project the input data into a latent space to group samples of the same class and distance them from other classes samples.
Relying on such geometrical separations, a simple linear classifier suffices to identify classes
but it is important to underline that such geometrical properties in the latent space
are \emph{implicitly learned}, i.e.,
the traditional training loss has no
knowledge of the latent space as it observes
points after the final linear layer, not in the latent space.

\begin{figure}[t]
\includegraphics[width=\columnwidth]{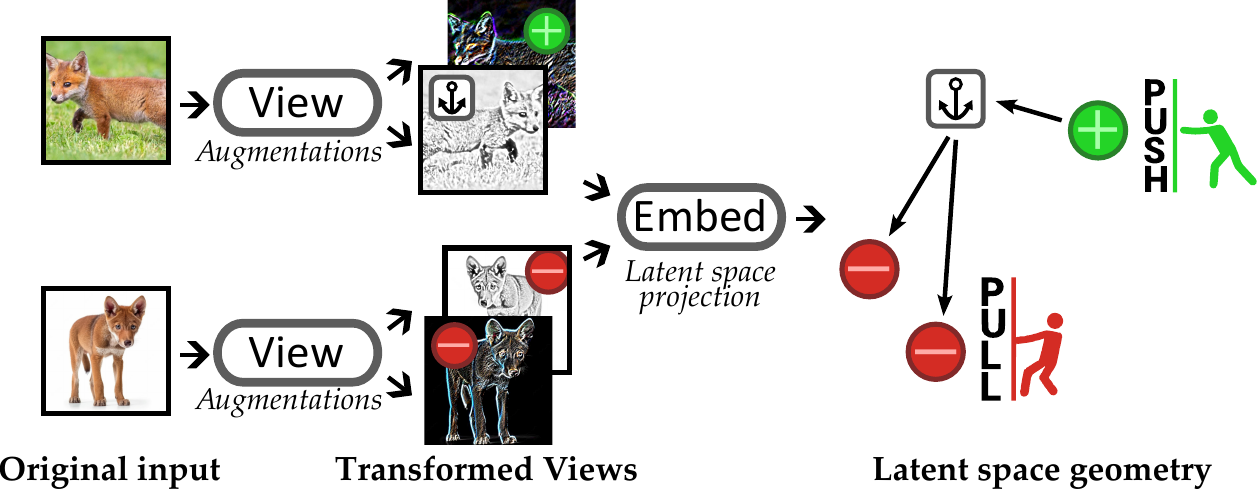}
\caption{Contrastive learning principles.
\label{fig:contrastive-learning-sketch}
}
\end{figure}

Conversely, contrastive learning is a special type of self-supervision
which aims to \emph{explicitly} enforce geometrical properties
in the latent space by means of data augmentations. Figure~\ref{fig:contrastive-learning-sketch}
sketches the principles behind the technique.
Input samples are transformed into ``views'' using
augmentation functions. Then views are compared using a ``contrastive game'':
a given view (an anchor) is compared in the latent space against other views of the same original image
(positive samples) and all other views (negative samples).
The contrastive loss function (namely InfoNCE for SimCLR) aims at pushing an anchor closer to its positives and distancing it from negatives.
This training is \emph{unsupervised} and a
given anchor is forced to be similar to other views of the same
image---in a sense, views of the same image form ``their own class''
even if negative samples can be of the same underlying class of the anchor.
Hence, this is a harder problem than when using supervision
and it is intentionally designed to push the learning of the representation.
Figure~\ref{fig:contrastive-learning-sketch} depicts
one specific configuration of (anchor, positive, negatives) but during
training all possible permutations are computed.
The figure also shows the setting defined by SimCLR, but other variations are possible (e.g., the anchor can be the original image, and some contrastive learning algorithms do not use negative samples~\cite{grill2020neurips-byol}).
Once the feature extractor (a.k.a., the representation) is pre-trained, it can be extended by fine tuning a linear layer and obtaining the final classifier. Overall, the more powerful the representation, the lower the number of samples required for fine-tuning.

\mypar{Target work}
Contrastive learning's appeal comes from (1) the ability to pre-train a feature extractor in an unsupervised manner and (2) the possibility of fine-tuning with just a few labeled samples. Specifically, \TARGETAUTHORS
pre-train using 100 unlabeled samples transformed based on packet time series transformations
(Change RTT and Time shift)\footnote{The augmentations used in \cite{horowicz2022imc-fewshotcl}
are inspired by the ones used in \cite{rezaei2019ICDM-ucdavis}.} using SimCLR and fine-tune using only up to 10 labeled samples.
Results show that, in some scenarios, classification performance is almost on-par with fully supervised training.

\begin{table*}[t]
    \centering
    \scriptsize
   \caption{Outline of the \TARGETPAPER contributions, their points of improvement and our contributions.}
    \label{tab:contributions_comparison}
    \begin{tabular}{
        @{}
        c@{$\,\,\,\,$} 
        c@{$\,\,\,\,$} 
        c@{$\,\,\,\,$} 
        @{}
    }
    \toprule
         \textbf{$\fontsize{7}{10}\selectfont\mathsf{\textsc{Ref-Paper}}$ contributions} & \textbf{Points of improvement} & \textbf{Our contribution} \\
         \midrule
         Test on 7 data augmentations on \UCDAVIS and \ISCX & 
            Issues with \ISCX; no statistical analysis of classif. perf.& 
            Extra 3 datasets; added statistical analysis  \\
         Evaluate SimCLR and its sensitivity to fine-tuning dataset size & 
            Lower performance with respect to the $\fontsize{7}{10}\selectfont\mathsf{\textsc{Ref-Paper}}$ &
            Expansion of training set size \\
         Performance comparison on fine-tuning with alternative inputs & 
            Analysis expansion to other parameters & 
            Fine-tuning sensitivity to dropout and augmentation \\
         \bottomrule
    \end{tabular}
\end{table*}

\mypar{Related work}
The closest related work to the \TARGETPAPER is \cite{towhid-netsoft22-boyl},
where the authors applied another off-the-shelf contrastive learning
method (Bootstrap Your Own Latent - BYOL \cite{grill2020neurips-byol} which, unlike SimCLR, does not rely on negative samples) by pre-training on augmented data and fine-tuning on a few samples. The authors relied on the same dataset as in the \TARGETPAPER but adopted packet time series as their input (rather than flowpics), leveraging the data transformations proposed by \cite{rezaei2019ICDM-ucdavis} and a ResNet18 architecture.
Overall, \cite{towhid-netsoft22-boyl} shows comparable performance with respect to the \TARGETPAPER.
A few more recent studies investigated contrastive learning on raw packet bytes as input~\cite{meng-kdd22,ziyi-networking22-cl-etc} and compared it against transfer learning and meta-learning \cite{guarino2023tma}, highlighting its recent relevance.

\section{Methodology
\label{sec:methodology}}

\subsection{Experimental goals}
As previously stated, this work aims (i) to \textit{reproduce} the main results of the \TARGETPAPER, employing the same methodologies and data, as well as (ii) to \textit{replicate} the \TARGETPAPER's results by considering three other datasets to confirm the validity of its findings. In more detail, our specific goals are:

\begin{enumerate}
\item[\bf (G0)] Provide a baseline analysis of the TC problem using classic Machine Learning (ML) models.
\item[\bf (G1)] Reproduce the benchmarking of the proposed data augmentation techniques for flowpics as a preliminary selection step for their use in contrastive learning. We aim for a quantitative reproduction {\bf (G1.1)} and a qualitative reproduction {\bf (G1.2)} of the augmentations ranks.
\item[\bf (G2)] Reproduce the SimCLR-related results with special attention to their
internal details (e.g., use of dropout,  projection layer size, training set size and impact of the combination of augmentations used).
\item[\bf (G3)] Replicate the benchmarking of data augmentation techniques on another set of public datasets.
\end{enumerate}

By addressing \textbf{G0} we aim to quantify the classification task's degree of difficulty using ML methods
and to assess if the use of more sophisticated modeling techniques is justified.
Additionally, since the \TARGETPAPER does not provide confidence intervals nor
perform any kind of statistical analysis of their results,
\textbf{G1} and \textbf{G2} aim to not only merely reproduce the original results but, also, to add a layer of statistical significance to them. Finally,
with \textbf{G3} we assess whether (statistically relevant) findings can be extended to other datasets.

\subsection{Experimental protocol
\label{sec:experimentalprotocol}
}
We closely followed the configurations and scenarios from the \TARGETPAPER which we complemented with ablation studies and modeling
campaigns. In this section we provide a summary of the main aspects of \cite{horowicz2022imc-fewshotcl}.
Details are deferred to the related evaluation sections where quote blocks as
\begin{quotethepaper}
\it\footnotesize example quote block
\end{quotethepaper}
\noindent highlight details from the \TARGETPAPER that require discussion.

\mypar{DL architectures}\label{sec:flowpic:architecture}
We adopted the same
CNN-based networks of the \TARGETPAPER,
namely a LeNet5~\cite{lenet5} (i.e., a \emph{mini}-flowpic)
and a larger version of it (i.e., a \emph{full}-flowpic).\footnote{The terminology in the \TARGETPAPER is overloaded as mini- and full-flowpic also refer to the resolution of the flowpic created.}
We also explored the impact of dropout layers and the size of SimCRL projection layers (see~\ref{sec:dropout-and-projection}).
In addition to our code artifacts, the printout of the networks is reported in App.~\ref{app:architectures}.

\mypar{Data augmentation}\label{sec:flowpic:dataaugment}
Next to applying no augmentation, we adopted
the 6 augmentations used in the \TARGETPAPER---3 packet time series transformations
(Change RTT, Time Shift and Packet Loss)
and 3 image transformations
(Rotation, Horizontal Flip, and Color jitter)---with the same
hyper-parameters (see \cite{horowicz2022imc-fewshotcl} for details).

\mypar{Training steps.} 
As in the \TARGETPAPER, we compared
two DL modeling techniques: \emph{fully supervised}
training and \emph{SimCLR + fewshot fine-tune} training.
For the former, samples are augmented before starting the training.
For the latter, given a labeled dataset and a selected
augmentation function, each sample is processed
to create 2 views of it using the Change RTT and Time shift transformations.
Both views are created when forming the mini batches used during training.
First, a representation of the dataset
is learned by \emph{pre-training} a model via SimCLR, contrasting pairs of augmented ``views''
of a sample.
Then, a new model is
formed by freezing the pre-trained representation and
combining it with a classifier layer which is
fine-tuned based on a few labeled samples.
As in the \TARGETPAPER,
we use Change RTT and Time Shift as data augmentation functions,
yet we complement the analysis testing other augmentation pairs too.
Augmentations are used only during pre-training.

\mypar{Comparing contributions}\label{sec:flowpic:context}
To compare our study with \cite{horowicz2022imc-fewshotcl}, we provide a summary of the \TARGETPAPER contributions in Table~\ref{tab:contributions_comparison}, highlighting the points of improvement we identified (some of them described in the coming sections) and the actions we took to expand on them, apart from the basic reproducibility and replicability efforts described above.
We remark that this table does not cover every contribution in this paper (e.g., it does not mention the added ML baseline), but rather contrasts what the original paper covered and how we increased the scope of the original contributions.

\subsection{System and Artifacts}
We performed 13 modeling campaigns, each consisting of the application of a target
configuration across multiple random seeds and data splits
(see Sec.~\ref{sec:evaluation} for details) for a total of 2,760
individual experiments. This entailed the
implementation of a modeling framework able
to properly track hyper-parameters,
performance metrics and other configurations
as well as output artifacts (e.g., models,
summary report, logs). For this tracking we relied on \texttt{AimStack}\footnote{\url{https://github.com/aimhubio/aim}} which we complemented based on our needs
(e.g., the framework has only minimal support for tracking output files). Both our modeling framework and the whole modeling
campaign outputs are provided as artifacts (see App.~\ref{app:artifacts}).

All experiments were run on Linux servers equipped with
multiple nVIDIA Tesla V100. Individual modeling experiment
duration span from a few minutes to hours and the overall
set of campaigns takes multiple weeks to run even in a distributed
setting.

\subsection{Datasets
\label{sec:datasets}
}

To address our goals we used
the four datasets summarized in Table \ref{tab:datasets}.
\UCDAVIS is used in the \TARGETPAPER while we selected
the others because of their interesting and complementary properties
with respect to \UCDAVIS:
($i$) they are collected in similar setups---research projects
related to mobile traffic monitoring---
($ii$) they cover a larger number of classes and users
behavior---\MIRAGEA and \UTMOBILENET are gathered from volunteering students
interacting with instrumented phones while \MIRAGEB focuses on
video meeting services---and ($iii$) they are imbalanced---the
$\rho$ values in the table reflect the ratio between
the number of samples of the largest and smallest class in a dataset;
notice the larger imbalance of the three datasets compared to \UCDAVIS,
which is an expected property of network traffic.
More importantly, all these datasets provide per-packet time series for the whole flows duration,
which is a key requirement for composing flowpic representations.
For instance, we cannot use the larger \APPCLASSNET~\cite{wang2022ccr-appclassnet}
and \CESNETTLS~\cite{luxemburk2023comnet-cesnettls} datasets
because they only provide the packet time series
for the first 20-30 packets of each flow.

\mypar{Data curation}
Each dataset is a collection of files (in either CSV or JSON format)
which we reprocessed into ``monolithic'' parquet files (a
well known serialization format used in data science)
encoding packet time series as \texttt{numpy} arrays.

\begin{table}[t]
    \centering
    \scriptsize
    \caption{Summary of datasets properties.}
    \label{tab:datasets}
\begin{tabular}{
    @{}l 
    @{$\,$}c 
    @{$\,$}c 
    @{$\,$}c
    @{$\,$}r
    @{$\,\,$}r 
    @{$\,\,$}r 
    @{$\,\,$}c
    @{$\,\,$}r
    @{}}
\toprule
\multirow{2}{*}{\bf Name}
& \multirow{2}{*}{\bf Partition}
& \multirow{2}{*}{\bf Filter}
& \multirow{2}{*}{\bf Classes}
& \multicolumn{4}{c}{\bf Flows}
& \bf Pkts
\\
&&&
& \emph{all}
& \emph{min} 
& \emph{max}
& \emph{$\rho$}
& \emph{mean}
\\
\midrule
\multirow{3}{*}{\cite{rezaei2019ICDM-ucdavis}~\UCDAVIS}
    & pretraining     & \multirow{3}{*}{\emph{none}}
                      & \multirow{3}{*}{5}   &      6,439 &        592 &      1,915   &        3.2  &      6,653\\
    & \HUMAN          & &                    &         83 &         15 &         20   &        1.3  &      7,666\\
    & \SCRIPT         & &                    &        150 &         30 &         30   &        1.0  &      7,131
\\
\midrule
\multirow{2}{*}{\cite{aceto2019mirage}~\MIRAGEA}
    & \multirow{2}{*}{\emph{n.a.}}
    & \emph{none}
    &  \multirow{2}{*}{20$^{(*)}$}
                              &    122,007 &      1,986 &     11,737 &        5.9 &         23\\
    &  & $>$\emph{10pkts} &   &     64,172 &      1,013 &      7,505 &        7.4 &         17
\\
\midrule
\multirow{3}{*}{\cite{guarino2021classification-mirage22}~\MIRAGEB}
     & \multirow{3}{*}{\emph{n.a.}}
     & \emph{none}
     & \multirow{3}{*}{9}
                             &     59,071 &      2,252 &     18,882 &        8.4 &      3,068\\
  & & $>$\emph{10pkts}&      &     26,773 &        970 &      4,437 &        4.6 &      6,598\\
  & & $>$\emph{1,000pkts}&      &      4,569 &        190 &      2,220 &       11.7 &     38,321
\\
 \midrule
 \multirow{2}{*}{\cite{utmobilenet21}~\UTMOBILENET}
    & \multirow{2}{*}{\emph{4-into-1}}
     & \emph{none}
     & 17
                             &     34,378 &        159 &      5,591 &       35.2 &        664\\
     & & $>$\emph{10pkts}
     & 14                    &      9,460 &        130 &      2,496 &       19.2 &      2,366
\\
\bottomrule
\end{tabular}
\\
\raggedright
$\rho$ : ratio between max and min number of flows---the larger the value, the higher the class imbalance; 
(*) Despite being advertised of having traffic from 40 apps, the \emph{public} version of the dataset 
only contains 20 apps.
\end{table}

As detailed in the table, \UCDAVIS is pre-partitioned
(and pre-filtered) by the authors of the dataset to create a large set of samples
for unsupervised training (namely pretraining) and two smaller testing set partitions,
namely \SCRIPT and \HUMAN.\footnote{According to \cite{rezaei2019ICDM-ucdavis},
both pretraining and \SCRIPT correspond to automated collection of data, while
\HUMAN is gathered monitoring traffic when real users were interacting
with the selected 5 services.} As such, we found no need to
alter the dataset beside the mere conversion to parquet.

Conversely, for the other three datasets we filtered out flows with
less than 10 packets and removed classes with less than 100 samples.
To replicate the setting provided in \UCDAVIS,
for \MIRAGEA and \MIRAGEB we also first
removed TCP ACK packets from time series and then discarded flows related
to background traffic.\footnote{Traffic is collected on mobile phones
with labeling ground-truth provided by \texttt{netstat}.
One measurement experiment generates traffic logs for a specific
target app. We processed such logs so that traffic of apps and services
different from the target app (e.g., netd deamon, SSDP, Android gms) is removed as it represents ``background'' traffic.}
We also highlight that \UTMOBILENET authors split the dataset into 4 partitions
(``Action-Specific'', ``Deterministic Automated``,   ``Randomized Automated'' and ``Wild Test'')
but we collated them into one.

The right-most column of the table details the average number of packets
in a flow. Notice how \UCDAVIS has very long flows while \MIRAGEA
is the dataset with shortest ones.
To further focus on very long flows, we also created a version
of \MIRAGEB with flows having more than 1,000 packets.

Lastly, through our curation we also created reference train/test splits for the datasets.
Specifically, since in the \TARGETPAPER the training dataset needs to have 100 samples,
for \UCDAVIS we create 5 folds (the smallest class in the dataset has 592 flows)
of 100 samples per-class each.
However, for the other datasets we opted for having 5 random splits each having
a random selection of 80\% of samples for training (and the rest for testing).
To ease replicability, we contribute the code used for our curation (which can be applied directly on original version of each dataset)
as well as our curated parquet files (see App.~\ref{app:artifacts}).

\mypar{Reproducibility}
Beside \UCDAVIS, the \TARGETPAPER also considers the \ISCXVPN and \ISCXTOR datasets,
but we discarded them after some preliminary investigations.
In fact, as acknowledged by \TARGETAUTHORS and
as well known in the literature, these datasets (even when combined)
contain only tens of viable flows for the analysis.
Hence, to use them, one would need to create multiple 15s windows
from the same flow to reach the 100 samples required for training,
which seems artificious.
More important, a recent work~\cite{jacobs2022ccs} carefully exposes
fallacies for these datasets which are rooted in some form of
data bias.\footnote{To be fair, the fallacies concern more the way the data bias can be unknowingly exploited to produced biased models. See ~\cite{jacobs2022ccs} for more details.}
While underlining these issues, we do not want to discredit the datasets but rather to justify our choice of discarding them from our study.

\mypar{Replicability}
Quantitatively reproducing research results on a dataset is a necessary starting point but not be the ultimate goal.
As we argued earlier, datasets age quickly in the TC field and new applications regularly emerge.
Thus, replication on novel datasets is equally important.
Qualitative agreement on a larger span of datasets brings the additional value of extending the validity of the findings.
For these reasons, we employ \MIRAGEA, \MIRAGEB and \UTMOBILENET to replicate insights
related to the comparison of data augmentation functions in the supervised setting.

\section{Evaluation
\label{sec:evaluation}
}

Unless differently stated, the results reported in this section
are collected using the \UCDAVIS dataset (training on the pretraining partition
and testing of the two predefined \HUMAN and \SCRIPT partitions).

\subsection{Providing a simple ML baseline (G0)}
\subsubsection{Approach}
We start with an ML baseline to assess to what extent DL techniques are justified---which would be the case if we observe a large discrepancy between ML and DL performance.

We used a classic XGBoost as our ML model, with default hyper-parameter values (100 estimators, max depth 6).
As input, we compared a mini-flowpic (a 32$\times$32 image flattened into a 1,024 values array) against the time series
of the packet size, direction and intertime of the the first 10 packets of a flow (i.e., 3 features of 10 values each all concatenated
into 30 elements arrays). We repeated the experiments 15 times and computed the 95\% confidence intervals using a t distribution.
Table~\ref{tab:G0} compares our results against those reported in the \TARGETPAPER for a LeNet5 CNN model trained without data augmentation
(but no confidence intervals are available).

\subsubsection{Results}
The trained forests have very short trees (an average depth of 1.7 for time series and 1.3 for flowpic input).
While trivial to execute, this analysis conveys interesting messages. For the \SCRIPT partition, ($i$) when using a flowpic representation, DL models have a slight advantage (about +2\%) over ML models; ($ii$) the advantage of flowpic over a simple  time series is more noticeable (about +4\%), which could be expected since the amount of information in an early time series (a few packets) is significantly smaller than what encoded in a flowpic (multiple seconds of traffic).

Instead, a different interpretation arises when considering the \HUMAN partition: ($i$) the results of ML are consistent with the observations in the \SCRIPT partition, i.e., using time series as input yields a score just a few percentage points lower than results using a flowpic input (6.74\% difference on average); however, ($ii$) the gap between DL and ML models when using flowpic is unexpectedly large (18.75\% on average).

\mypar{Takeaway}{\it
Based on the \TARGETPAPER results, our expectations were to have models offering
similar performance on both testing partitions. Yet, we observed a large discrepancy for \HUMAN
which calls for a deeper analysis that we carry out in the following sections.
}

\begin{table}[t]
    \footnotesize

 \caption{(G0) Baseline ML  performance without augmentations in a supervised setting.}
 \label{tab:G0}
 \begin{tabular}{lll ll}
  \toprule
\bf Input (size) &  \bf Model  &  \bf Origin &  \multicolumn{2}{c}{\bf Accuracy $\pm$ 95\%CI}\\
 \cmidrule{4-5}
 \multicolumn{3}{c}{} &  \SCRIPT  & \HUMAN \\
 \midrule
flowpic $(32\times 32)$ &   CNN LeNet5 & \      \cite{horowicz2022imc-fewshotcl} & 98.67        &          92.40    \\ 
flowpic $(32\times 32)$  &   XGBoost & ours &  96.80\tinytiny{0.37} &  73.65\tinytiny{2.14}\\
time series $(3\times 10)$ &  XGBoost & ours & 94.53\tinytiny{0.56} &   66.91\tinytiny{1.40}\\
 \bottomrule 
 \end{tabular}
 \\
 \raggedright
Each \emph{ours} is an aggregations of 15 experiments (5 splits $\times$ 3 seeds).
 \end{table}

\subsection{Reproducing quantitative results of data augmentation (G1.1)}\label{sec:G1.1}

%
%
\begin{table*}

\caption{Comparing data augmentation functions in a supervised training. 
Values marked as ``ours'' correspond to the average accuracy
across 15 modeling experiments and the related 95-th confidence intervals.
\label{tab:augmentation-at-loading}
}

\scriptsize
\begin{tabular}{
    @{}l
    @{$\,\,\,$}r
    @{$\,\,\,$}r
    @{$\,\,\,$}r
    r@{$\,$}r@{$\,\,\,$}
    r@{$\,$}r@{$\,\,\,$}
    r@{$\,$}r
    r@{$\,\,\,$}
    r@{$\,\,\,$}
    r
    r@{$\,$}r@{$\,\,\,$}
    r@{$\,$}r@{$\,\,\,$}
    r@{$\,$}r
    r@{$\,$}r@{$\,\,\,$}
    r@{$\,$}r@{$\,\,\,$}
    r@{$\,$}
    r@{}
}
\toprule
& \multicolumn{9}{c}{\bf Test on \SCRIPT}
& \multicolumn{9}{c}{\bf Test on \HUMAN}
& \multicolumn{6}{c}{\bf Test on \LEFTOVER$^{\,\dagger}$}
\\
\cmidrule(r){2-10}
\cmidrule(r){11-19}
\cmidrule(r){20-25}
& \multicolumn{3}{c}{{\it from} \cite{horowicz2022imc-fewshotcl}}
& \multicolumn{6}{c}{\it ours}
& \multicolumn{3}{c}{{\it from} \cite{horowicz2022imc-fewshotcl}}
& \multicolumn{6}{c}{\it ours}
& \multicolumn{6}{c}{\it ours}
\\
\cmidrule(r){2-4}
\cmidrule(r){5-10}
\cmidrule(r){11-13}
\cmidrule(r){14-19}
\cmidrule(r){20-25}
\multicolumn{1}{r}{\it flowpic res}
    & \multicolumn{1}{c}{32}    
    & \multicolumn{1}{c}{64}    
    & \multicolumn{1}{c}{1500}      
    & \multicolumn{2}{c}{32}  
    & \multicolumn{2}{c}{64}  
    & \multicolumn{2}{c}{1500}
    & \multicolumn{1}{c}{32}    
    & \multicolumn{1}{c}{64}    
    & \multicolumn{1}{c}{1500}      
    & \multicolumn{2}{c}{32}  
    & \multicolumn{2}{c}{64}  
    & \multicolumn{2}{c}{1500}
    & \multicolumn{2}{c}{32}  
    & \multicolumn{2}{c}{64}  
    & \multicolumn{2}{c}{1500}
\\
\cmidrule(r){1-1}
\cmidrule(r){2-4}
\cmidrule(r){2-4}
\cmidrule(r){5-10}
\cmidrule(r){11-13}
\cmidrule(r){14-19}
\cmidrule(r){20-25}
No augmentation      & 98.67 & 99.10 & 96.22    & 95.64 & \tinytiny{ 0.37}  & 95.87 & \tinytiny{ 0.29}  & 94.93 & \tinytiny{ 0.72}  
                     & 92.40 & 85.60 & 73.30    & 68.84 & \tinytiny{ 1.45}  & 69.08 & \tinytiny{ 1.35}  & 69.32 & \tinytiny{ 1.63}  
                                                & 95.78 & \tinytiny{ 0.29}  & 96.09 & \tinytiny{ 0.38}  & 95.79 & \tinytiny{ 0.51}  
\\
Rotate               & 98.60 & 98.87 & 94.89    & 96.31 & \tinytiny{ 0.44}  & 96.93 & \tinytiny{ 0.46}  & 95.69 & \tinytiny{ 0.39}  
                     & 93.73 & 87.07 & 77.30    & 71.65 & \tinytiny{ 1.98}  & 71.08 & \tinytiny{ 1.51}  & 68.19 & \tinytiny{ 0.97}  
                                                & 96.74 & \tinytiny{ 0.35}  & 97.00 & \tinytiny{ 0.38}  & 95.79 & \tinytiny{ 0.31}  
\\
Horizontal flip      & 98.93 & 99.27 & 97.33    & 95.47 & \tinytiny{ 0.45}  & 96.00 & \tinytiny{ 0.59}  & 94.89 & \tinytiny{ 0.79}  
                     & 94.67 & 79.33 & 87.90    & 69.40 & \tinytiny{ 1.63}  & 70.52 & \tinytiny{ 2.03}  & 73.90 & \tinytiny{ 1.06}  
                                                & 95.68 & \tinytiny{ 0.40}  & 96.32 & \tinytiny{ 0.59}  & 95.97 & \tinytiny{ 0.80}  
\\
Color jitter         & 96.73 & 96.40 & 94.00    & 97.56 & \tinytiny{ 0.55}  & 97.16 & \tinytiny{ 0.62}  & 94.93 & \tinytiny{ 0.68}  
                     & 82.93 & 74.93 & 68.00    & 68.43 & \tinytiny{ 2.82}  & 70.20 & \tinytiny{ 1.99}  & 69.08 & \tinytiny{ 1.72}  
                                                & 96.93 & \tinytiny{ 0.56}  & 96.46 & \tinytiny{ 0.46}  & 95.47 & \tinytiny{ 0.49}  
\\
Packet loss          & 98.73 & 99.60 & 96.22    & 96.89 & \tinytiny{ 0.52}  & 96.84 & \tinytiny{ 0.63}  & 95.96 & \tinytiny{ 0.51}  
                     & 90.93 & 85.60 & 84.00    & 70.68 & \tinytiny{ 1.35}  & 71.33 & \tinytiny{ 1.45}  & 71.08 & \tinytiny{ 1.13}  
                                                & 96.99 & \tinytiny{ 0.39}  & 97.25 & \tinytiny{ 0.39}  & 96.84 & \tinytiny{ 0.49}  
\\
Time shift           & 99.13 & 99.53 & 97.56    & 96.71 & \tinytiny{ 0.60}  & 97.16 & \tinytiny{ 0.49}  & 96.89 & \tinytiny{ 0.27}  
                     & 92.80 & 87.33 & 77.30    & 70.36 & \tinytiny{ 1.63}  & 71.89 & \tinytiny{ 1.59}  & 71.08 & \tinytiny{ 1.33}  
                                                & 97.02 & \tinytiny{ 0.50}  & 97.51 & \tinytiny{ 0.46}  & 97.67 & \tinytiny{ 0.29}  
\\
Change RTT           & 99.40 & 100.00 & 98.44    & 97.29 & \tinytiny{ 0.35}  & 97.02 & \tinytiny{ 0.46}  & 96.93 & \tinytiny{ 0.31}  
                     & 96.40 & 88.60 & 90.70    & 70.76 & \tinytiny{ 1.99}  & 71.49 & \tinytiny{ 1.59}  & 71.97 & \tinytiny{ 1.08}  
                                                & 98.38 & \tinytiny{ 0.18}  & 97.97 & \tinytiny{ 0.39}  & 98.19 & \tinytiny{ 0.22}  
\\
\cmidrule(r){1-1}
\cmidrule(r){2-4}
\cmidrule(r){2-4}
\cmidrule(r){5-10}
\cmidrule(r){11-13}
\cmidrule(r){14-19}
\cmidrule(r){20-25}
\multicolumn{1}{r}{\it mean diff$^{\,\,\ddagger}$}
    & \multicolumn{3}{c}{}
    & \multicolumn{2}{c}{-2.05}
    & \multicolumn{2}{c}{-2.26}
    & \multicolumn{2}{c}{-0.63}
    & \multicolumn{3}{c}{}
    & \multicolumn{2}{c}{-21.96}
    & \multicolumn{2}{c}{-13.27}
    & \multicolumn{2}{c}{-9.13}
\\
\bottomrule
\end{tabular}
\\
\raggedright
Each of \emph{our} result is an aggregation of 15 experiments (5 splits $\times$ 3 seeds).
\\
$\dagger$ We named ``leftover'' the samples from the \emph{pretraining} partition not belonging to the 100 samples 
of a given split. Traditionally this would correspond to the test set.\\
$\ddagger$ \emph{mean diff} corresponds to the difference between our assessment and the expected value averaged across
augmentations for each given flowpic resolution.
\end{table*}

We continue by reproducing results related to Tables 1--2 of
\cite{horowicz2022imc-fewshotcl}, which contrast different augmentations
applied in a supervised setting.

\subsubsection{Approach} Given the unexpected results of the ML baseline,  we adopted a very careful approach, that we detail in what follows. \TARGETAUTHORS wrote:
\begin{quotethepaper}
\it\footnotesize
For all experiments, for training set we use only 100 ``triggered by script''
flows per class, and for test set we follow the
experiments by [16] randomly choosing 30 flows for each
class for a ``triggered by script'' test set and 15 flows per class
for "triggered by human" test set.
[...] For all experiments, we apply each of the augmentations
10 times on the 100 samples per class training set, which
increase the training set to 1000 images per class.
We also train without any augmentation as baseline experiments and term it "no aug". For all experiments we allocated
20\% of the images for validation, and early stopped the training when the validation loss stopped improving
\end{quotethepaper}

First of all, recall that \UCDAVIS is composed of three partitions
explicitly named to express the intention of separating a portion
of the data used for pre-training from another reserved
for testing and fine-tuning (see Table~\ref{tab:datasets}).
Although the authors use  ``triggered by script'' twice,
we interpreted that 100 flows are selected for training
from the large pretraining partition, while using the remaining
two partitions (\SCRIPT and \HUMAN)\footnote{While \SCRIPT is perfectly balanced with 30 flows
per class, \HUMAN has three classes with 15 samples, and the
remaining two have 18 and 20 samples respectively. Given the very small
imbalance we considered irrelevant to resample the partitions to have exactly 15 samples per classes.
Thus, we use \SCRIPT and \HUMAN as is.} for testing and
fine-tuning.\footnote{Authors later clarified that
they combined pretraining and \SCRIPT. However this minor difference does
not affect the results of our investigation.
}

Secondly, we did not find explicit mentions of how many experiments
were performed to gather the results, nor do the tables report
confidence intervals.
Yet, we assume that several runs were carried out, as it
is common practice when performing modeling campaigns to assess
the performance across different dataset splits and models
initialization.\footnote{Authors did not provide us more details on this aspect.}
Since the original experiments were done by training with 100 samples per-class (and
the classes are imbalanced) doing
a traditional k-folds cross validation is not possible.
Thus, as from Sec.~\ref{sec:datasets},
we created $k$ splits by sampling without replacement groups
of 100 samples for each class from the pretraining partition.
Then, a given set of 100 samples is split randomly $s$ times,
with each split corresponding to a 80/20 train/validation split
for training.
Using these data, we performed a campaign to test
the 7 augmentations across k=5 splits each having s=3 train/validation splits
for a total of 105 experiments.
This is repeated for the three flowpic resolutions 
with the same training settings as in the \TARGETPAPER: static learning rate at 0.001, early stopping
on validation loss after 5 steps in which the loss
does not improve by more than 0.001, batch size of 32, performance
measured via accuracy, flowpic created from the first 15s of a flow.

\subsubsection{Results}
Table~\ref{tab:augmentation-at-loading} summarizes  our results reporting the mean accuracy and related 95\% CI for each scenario. To ease their comparison, we copy the reference results from the \TARGETPAPER and summarize in the last row the differences across scenarios with a simple arithmetic mean. We complement the evaluation of the \TARGETPAPER by reporting a new test set corresponding to all pretraining samples not belonging to a selected 100 samples split (i.e., what would be called a test set in a traditional evaluation). As, to the best of our understanding, these samples have been discarded in the \TARGETPAPER, we refer to this test set as \LEFTOVER.

Overall, we obtained lower performance
than what was previously reported. While
differences are modest on
\SCRIPT, we observe a reduction of over $20\%$  on \HUMAN---this is coherent with what we observed for the ML baseline.
Notice that no gap appears when comparing \SCRIPT with \LEFTOVER.

The gap is (slightly) reduced when using a higher resolution flowpic but the lower performance on \HUMAN (and the larger confidence intervals with respect to \SCRIPT and \LEFTOVER) suggests the presence of a hidden problem with this predefined test set.
Understanding the reason of this gap is important to
verify the validity of our study.
However,
we defer a significant portion of our study of the performance gap to App.~\ref{app:gap-dataset}--\ref{app:gap-reproduce} and we report only the salient aspects of our investigation in the following sections.

We highlight that, while for 32$\times$32 and 64$\times$64 experiments run in about 1 min,
it takes about 30min to run one experiment on 1500$\times$1500.
Given this computational cost, motivated by the marginal performance gap across resolutions and as done by
\TARGETAUTHORS, in the remainder of the paper we focus only on the 32$\times$32 resolution.

\begin{figure}[t]
    \centering
    \includegraphics[width=0.5\textwidth]{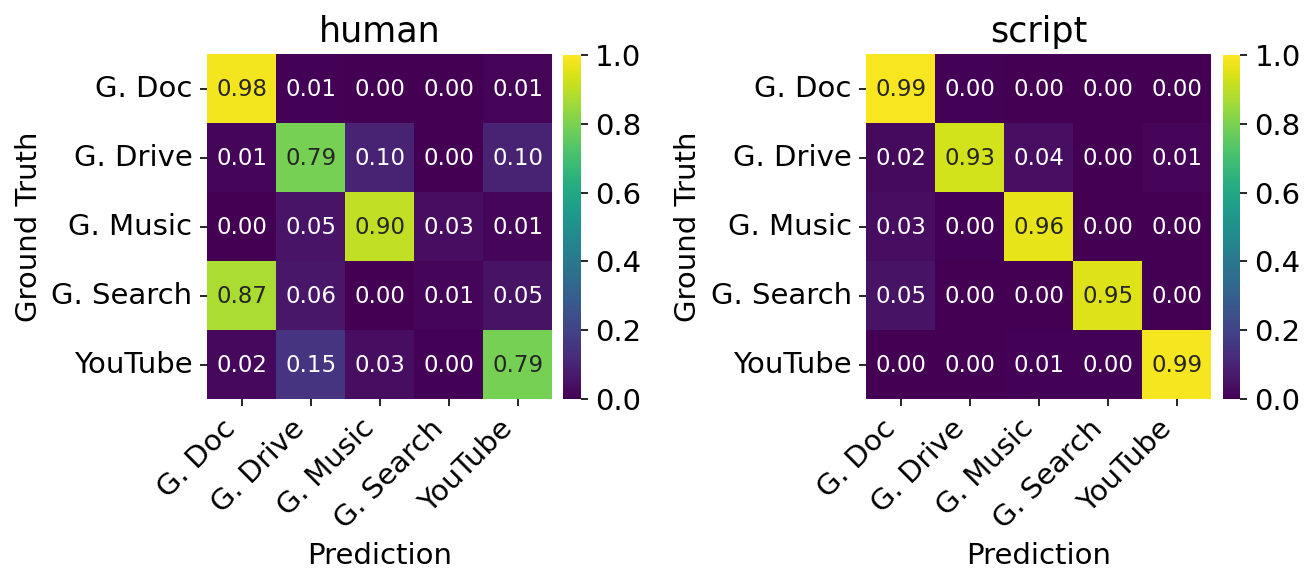}
    \caption{Average confusion matrixes for the 32$\times$32 resolution across all experiments in Table~\ref{tab:augmentation-at-loading}.}
    \label{fig:ucdavis-supervised-confusion-matrix}
\end{figure}

\subsubsection{Root cause of performance gap}
We reiterate that we do not apply any pre-processing (e.g., filtering, reshaping)
to the \UCDAVIS dataset beside consolidating  the original CSV files (one for each flow) into a monolithic parquet file. Thus, we conjectured that the root case of the performance problem might be rooted in the data itself.

To start verifying this assumption,  the heatmaps in Fig.~\ref{fig:ucdavis-supervised-confusion-matrix} break down the results in Table~\ref{tab:augmentation-at-loading} by showing the average per-class accuracy across the 105 runs for the 32x32 flowpic resolution. Specifically, we summed all the confusion matrices  for \SCRIPT and \HUMAN and we normalized them by row.
For \HUMAN we observe multiple sources of confusion with
\emph{Google doc} and \emph{Google search} having the most evident clash.
Conversely, no specific issues can be detected for \SCRIPT.

\begin{figure}
    \centering
    \includegraphics[width=\columnwidth]{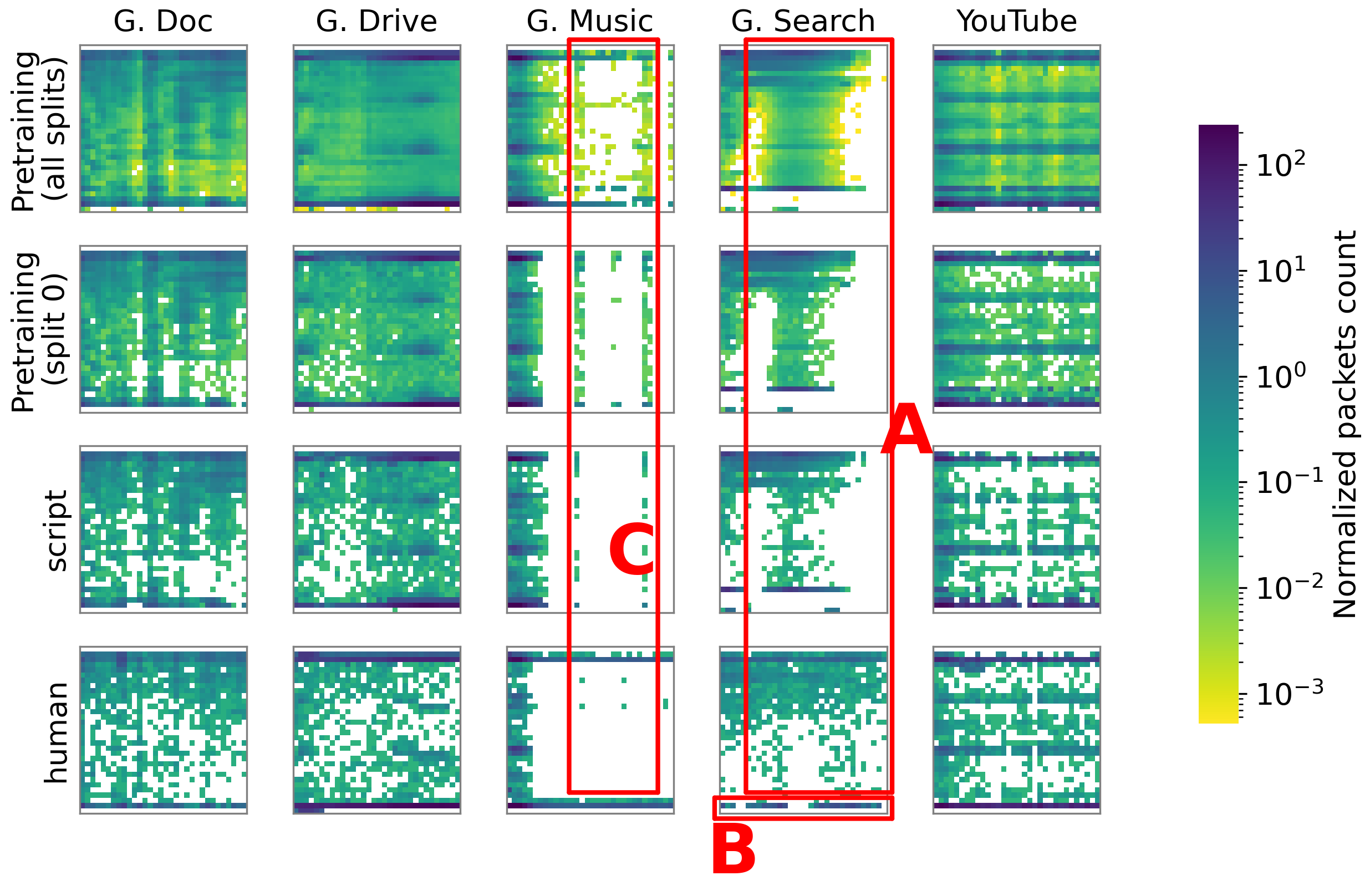}
    \caption{Average 32$\times$32 flowpic for each class across dataset partitions.}
    \label{fig:ucdavis-flowpic}
\end{figure}

To drill down, Fig.~\ref{fig:ucdavis-flowpic}
collects an average flowpic per class across the original dataset partitions
and one training split.
Recall that the horizontal axis of a flowpic corresponds to time
(time zero on the left) while the vertical axis corresponds to packet sizes (zero length on the top).

The first row in Fig.~\ref{fig:ucdavis-flowpic} corresponds to all flows available in the pretraining
partition, while the second one corresponds to
a training split, i.e., an aggregation of 100 samples per class.
We can clearly see that the reduction of samples
has a visual impact, but overall the first two rows
are visually very similar. The third and the fourth
rows correspond to the \SCRIPT and \HUMAN partition respectively, i.e.,
they have 30 and $\approx$15 samples per class. When
comparing the last two rows with the first two, we can
clearly see differences which we further annotate
with rectangles. Notice how \emph{Google search}
is expected to have two vertical groups of pixels
around the left-axis and the center of the picture.
Surprisingly, for \HUMAN these groups are ``shifted''
to the right (rectangle A). Moreover, notice
how all splits but \HUMAN saturate the
maximum packet size for \emph{Google search}---there is a distinctive horizontal line (around
pixels on row 28) for \HUMAN (rectangle B) while
in the other cases there are distinct dark lines at row 32.
Interestingly, Fig.~\ref{fig:ucdavis-flowpic}
also
highlights macroscopic differences  for \emph{Google music}---vertical ``stripes''
of pixels are visible in all splits but \HUMAN (rectangle C).
Yet, according to Fig.~\ref{fig:ucdavis-supervised-confusion-matrix},
this seems less of a problem.
We conjecture that this might be due to the stark difference
between \emph{Google music} and the other services.
In other words, despite the different behavior between
the partitions, \emph{Google music} is still very different
from the other 4 classes (thus it might be easier
to classify).

The analysis of the average flowpics supports
the idea of a data shift, of which
we provide further evidence in the Appendix.
Specifically, we support this statement by
($i$) adding more evaluations on \UCDAVIS (App.~\ref{app:gap-dataset}),
($ii$) resorting to content from \cite{rezaei2019ICDM-ucdavis} which introduced the \UCDAVIS dataset (App.~\ref{app:gap-ucdavis}), and ($iii$)  verifying code artifacts from \cite{rezaei2019ICDM-ucdavis} to help us rule out possible mistakes in our approach (App.~\ref{app:gap-reproduce}).
Summarizing this extensive material, the existence of a data shift is pointed out by both ($i$) and ($ii$) and we confirm ($iii$) as our verification yields expected results.

\mypar{Takeaway}{\it
Given the strong evidence provided by our analysis,
we concluded that the \HUMAN test split is affected
by a data-shift. Yet, we cannot  comment
on the reason why this was not detected in the \TARGETPAPER.\footnote{Authors did not provide us comments about this aspect.}}


\subsection{Reproducing qualitative ranking of data augmentation (G1.2)}\label{sec:G1.2}
\subsubsection{Approach} The original key question behind benchmarking the different
augmentations was to understand if, and by how much, they
were beneficial with respect to not performing any augmentation.
\TARGETAUTHORS wrote
\begin{quotethepaper}
\footnotesize\it
In all the nine experiments changing the RTT was the best
performing augmentation. The improvement varies from 1\%
for the QUIC script dataset (where the "no aug" accuracy
was already 98.7\%) up to 17.4\% improvement for the most
challenging dataset, the QUIC human.
\end{quotethepaper}
Without more details on the \TARGETPAPER it is very difficult to compare
against the reported results.
We opted instead for performing a statistical analysis of our
modeling campaign to understand if
Change RTT and Time shift were the best performing
augmentations as reported in the \TARGETPAPER.

The CI values in Table~\ref{tab:augmentation-at-loading} show clear
overlaps between different augmentations.
To investigate our results, we treat each augmentation as a different classifier and compare them according to the procedures presented in \cite{demvsar2006statistical}. First, accuracy results are turned into rankings (e.g., if augmentations A, B and C yield an accuracy of 0.9, 0.7 and 0.8, their associated rankings would be 1, 3, and 2) with ties being assigned with the average ranking of the group (e.g., if augmentations A, B and C yield 0.9, 0.9 and 0.8, their associated rankings would be 1.5, 1.5 and 3). This process is repeated across all tested datasets and splits. Then, an average ranking value is extracted per augmentation. These values are compared pairwise using a post-hoc Nemenyi test, which compares these average rankings to decide if the performance difference between augmentations is significant. This decision is made using a Critical Distance (CD) in ranking equal to $CD = q_{\alpha}\sqrt{\frac{k(k+1)}{6N}}$, where $q_{\alpha}$ is based on the Studentized range statistic divided by $\sqrt{2}$, $k$ is equal to the number of augmentations compared and $N$ is equal to the number of samples used.

\begin{figure}
    \centering
    \includegraphics[width=\linewidth]{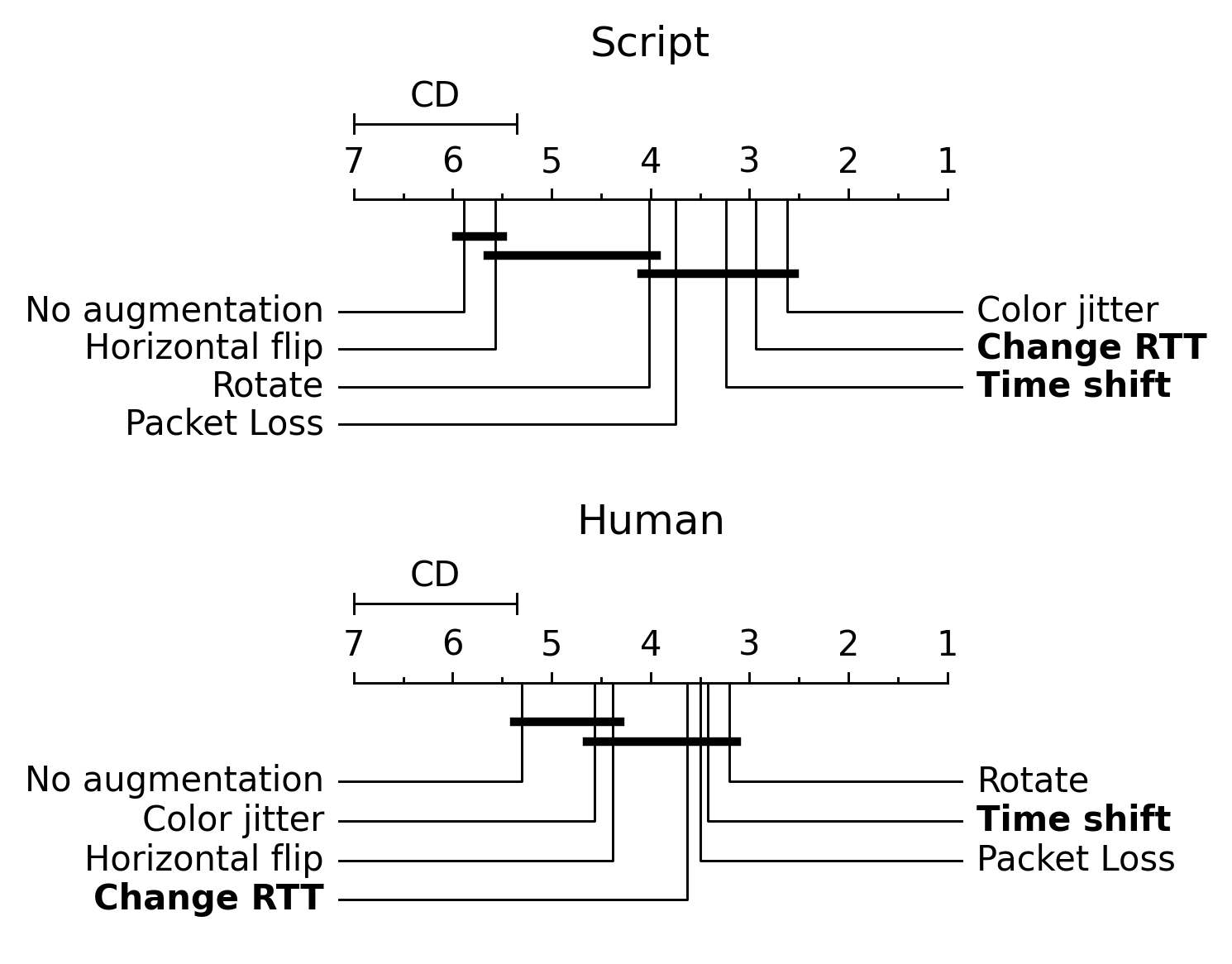}
    \caption{Critical distance plot of the accuracy obtained with each augmentation for the 32$\times$32 and 64$\times$64 resolutions. Augmentations joined by a horizontal line are not statistically different. The lower the ranking (closer to 1, the right side of the plot) the better the performance. Transformations highlighted in bold are selected as the best performing one in the \TARGETPAPER.}
    \label{fig:augmentation_ranks}
\end{figure}

\subsubsection{Results} Figure \ref{fig:augmentation_ranks} displays the results of these comparisons.
We combined the 32$\times$32 and 64$\times$64 resolutions as we did not find statistically significant differences between them (see App.~\ref{appendix:tukey}). In our case, with $\alpha = 0.05$, $k = 7$ and $N = 30$ and $q_{0.05} = 2.949$ the critical distance is $CD = 1.644$. The closer an augmentation is to the right side of the plot (a higher average rank), the better the performance.

From our analysis for the \SCRIPT partition, we cannot conclude significant differences within three groups, which we sort by increasing performance: {\{No augmentation and Horizontal flip\}; \{Horizontal flip and Rotate\}; \{Rotate, Packet loss, Time shift, Change RTT and Color jitter\}.  
Similar groups exist also for the \HUMAN partition. As annotated in Fig. \ref{fig:augmentation_ranks},  \TARGETAUTHORS selected  {\bf Change RTT }  and {\bf Time shift} as the best augmentations: whereas these augmentations are in the best performing group both for \SCRIPT and \HUMAN, it is easy to gather that other transformations consistently appear in the same (statistically relevant) group.

\mypar{Takeaway}{\it
On the one hand, the Time shift and Change RTT transformations are in the best performing group, a finding aligned with the ones in the \TARGETPAPER. On the other hand, from a statistical viewpoint, they are not distinguishable from other options, like Color jitter  (for \SCRIPT) or Rotate  (for \HUMAN) or Packet Loss (for both). }

\subsection{Reproducing constrastive learning results (G2)}\label{sec:G2}

\subsubsection{Approach} The second goal of our reproducibility study concerns the
use of contrastive learning and fine-tuning.
A few observations are needed to contextualize the
modeling campaign to perform.
\mypar{Augmentations for SimCLR}
First of all, we need to select augmentations  for SimCLR.
\TARGETAUTHORS wrote:
\begin{quotethepaper}
\footnotesize\it
we selected
to use 'Change RTT' by $\alpha \sim U$ [0.5, 1.5] together with Time
Shift by $b \sim U$ [-1,1]. In each training step, a double batch
of 32 unlabeled images (taken from the pool of 100 unlabelled samples per class) is loaded after applying the two
augmentations above.
\end{quotethepaper}
This confirms the traditional SimCLR approach where two views are obtained from each sample in a training mini-batch.
However, precisely how the transformations are applied is open to interpretation,
e.g., one after the other? If so, in which order? A separate transformation for each view?
These are design choices likely depending on the task at hand.
For instance, the original SimCLR paper~\cite{chen2020simple-simclr} shows that both which transformations are selected and the order in which they are chained are relevant decisions. Since a full ablation study on this aspect is well beyond our scope,  we opted for \emph{applying the two transformations in random order
for every image in a mini-batch}. Yet, given our ranking analysis showed equivalence among multiple top performing transformations,
we also perform a small-scale ablation study considering three other pairs beside the pair selected in the \TARGETPAPER.\footnote{\TARGETAUTHORS clarified with us that the two transformation were chained and to check their repository \cite{unofficialgit}. Yet, we reiterate that such repository cannot be used to reproduce the results of the \TARGETPAPER (see App~\ref{quote: icdm-datashit-explain}).
}

\mypar{Networks for SimCLR}
Even more subtle design choices relate to the application
of dropout and the projection layer size used in SimCLR.
\TARGETAUTHORS wrote:
\begin{quotethepaper}
\footnotesize\it
As depicted in Figures 6 and 7, our
architectures comprise seven layers, the ReLU activation
function is applied to the output of every convectional and
fully-connected layer and dropout with probabilities of 0.25
and 0.5 are used in order to reduce overfitting.
[...]
\end{quotethepaper}
The figures mentioned refer to the ``mini'' (for 32$\times$32 and 64$\times$64)
and ``full'' (for 1500$\times$1500) architectures. We underline that the mini architecture is
identical to the original LeNet5~\cite{lenet5}.
However, from the quote we identify a number of layers miscount,
as the full version has one layer less than the mini version.
In fact, the ``flatten'' layer is reported
only in the full diagram but is actually needed in both versions,
and the diagram clearly shows that the full version has
one less fully connected layer than the mini version
(see Fig. 6-7 in \cite{horowicz2022imc-fewshotcl}).
Moreover, based on the quote it is not clear if
dropout was applied to both architectures or just the full
version (as the original LeNet5 does not rely on dropout).
Given the lower resolution, dropout
might be not needed for 32$\times$32.

Regarding the SimCLR projection, \TARGETAUTHORS wrote:
\begin{quotethepaper}
\footnotesize\it
For the representation extractor $f(\cdot)$ we employed the 5
first layers of the CNN architectures described in A.1 and
replaced the last 2 layers with 2 linear layers sized 120 and
30. Thus, resulting with a 120 dimensional representation
vector $h = f(flowpic)$ and $z = g(h)$ dimensional similarity
vector.
\end{quotethepaper}
This refers to what is known as the \emph{projection layer} of
the feature extractor. In a nutshell, and based on our interpretation
of the quote, after the convolutional blocks,
the network have a 120-120-30 series of linear layers.
However, since the supervised network was using a latent space of size 84,
we investigated networks considering both 30 and 84 as final projection
layer dimension.

An assessment of these lower level details can be key
to obtain a fair comparison against the performance reported in the \TARGETPAPER.
For reference, we report the listing of the architecture used in App.~\ref{app:architectures}.

\begin{table}[t]
\centering
\footnotesize
\caption{
Impact of dropout and SimCLR projection layer dimension on fine-tuning  (32$\times$32 only, with 10 samples for fine-tuning training).
\label{tab:dropout-projection-layer}}
\centering
\begin{tabular}
{
    c
    r@{$\,$}l
    r@{$\,$}l
    r@{$\,$}l
    r@{$\,$}l
    r@{$\,$}l
    r@{$\,$}l
    r@{$\,$}l
    r@{$\,$}l
}
\toprule 
& \multicolumn{4}{c}{\it test on \SCRIPT} 
& \multicolumn{4}{c}{\it test on \HUMAN}
\\
\cmidrule(r){1-1}
\cmidrule(r){2-5}
\cmidrule(r){6-9}
Proj. dim
& \multicolumn{2}{c}{w/ dropout}
& \multicolumn{2}{c}{w/o dropout}
& \multicolumn{2}{c}{w/ dropout}
& \multicolumn{2}{c}{w/o dropout}
\\
\cmidrule(r){1-1}
\cmidrule(r){2-5}
\cmidrule(r){6-9}
30      & 91.81&\tinytiny{ 0.38}\reddagger      & 92.18&\tinytiny{ 0.31}      & 72.12&\tinytiny{ 1.37}\redddagger      & 74.69&\tinytiny{ 1.13}     
\\
84      & 92.02&\tinytiny{ 0.36}                & 92.54&\tinytiny{ 0.33}      & 73.31&\tinytiny{ 1.04}                 & 74.35&\tinytiny{ 1.38}     
\\
\bottomrule
\end{tabular}
\\
\raggedright
\footnotesize
Each value is an aggr. of 125 exp. (5 splits $\times$ 5 SimCRL seeds $\times$ 5 fine-tune seeds);
The reference value for \textcolor{red}{$\dagger$} from \cite{horowicz2022imc-fewshotcl} reports in the text (94.5\% for 10 samples); for \textcolor{red}{$\ddagger$} no
specific values are reported but should be $\approx$80\% based on Fig. 4 of \cite{horowicz2022imc-fewshotcl}.
\end{table}

\subsubsection{Results\label{sec:dropout-and-projection}}

As before, we follow the parameters
described in the \TARGETPAPER, namely
batch size of 32, patience of 3 on the top-5 accuracy when training
with SimCLR (temperature=0.07, learning rate=0.001) and patience
of 5 on train (min delta=0.001) during fine-tuning (learning rate=0.01).

Table~\ref{tab:dropout-projection-layer} details the results
of our ablation campaign to understand the impact of  dropout
and the projection layer.\footnote{We did also an ablation of dropout
before running results for Table~\ref{tab:augmentation-at-loading}. Details are reported
in Appendix~\ref{app:dropout-in-supervised}. The takeaway is that even for 1500$\times$1500
resolution there are minimal differences introduced by dropout.
Yet, results in Table~\ref{tab:augmentation-at-loading} reflect the use of dropout
as intended in the original study.
}
Each value in the table corresponds
to the mean and related 95-th percentiles CI
across 125 experiments and fine-tuning using 10 training samples.
As we expected,
we observe poorer performance when
testing on \HUMAN, while performance on \SCRIPT is just a few points
lower than for supervised training.
When considering a projection layer of 30 units, we can
observe that dropout does not provide a significant difference
for \SCRIPT; conversely, removing dropout makes a stark difference
when testing on \HUMAN. Increasing the projection layer dimension
does not provide a significant gain.
We conclude than that we can rely on a network \emph{without} dropout
(differently from the \TARGETPAPER)
but we confirm the original choice of a projection layer of 30 units.

\begin{table}[t]
\scriptsize
\centering
\caption{Comparing the fine-tuning performance when using different
pairs of augmentations for pretraining (32$\times$32 resolution, 
fine-tuning on 10 samples only).}
\label{tab:augmentation-pairs}
\begin{tabular}{
    @{}
    r 
    @{$\:$}c
    @{$\:$}c
    @{$\:$}c
    @{$\:$}c
    @{$\:$}c
    @{$\:$}c
    @{}
}
\toprule
\multicolumn{1}{@{}r}{\it 1st augment.}    
    & Change RTT\redasterisc 
    & \multicolumn{2}{c}{Packet loss}               
    & \multicolumn{2}{c}{Change RTT}                 
    & Color Jitter         
\\
\cmidrule(r){2-2}
\cmidrule(r){3-4}
\cmidrule(r){5-6}
\cmidrule(r){7-7}
\multicolumn{1}{@{}r}{\it 2nd augment.}    
    & Time shift\redasterisc & Color jitter          
    & Rotate                
    & Color Jitter          
    & Rotate                
    & Rotate                
\\
\cmidrule(r){2-2}
\cmidrule(r){3-4}
\cmidrule(r){5-6}
\cmidrule(r){7-7}
\it test on \SCRIPT    & 92.18\tinytiny{ 0.31}    & 90.17\tinytiny{ 0.41}    & 91.94\tinytiny{ 0.30}    & 91.72\tinytiny{ 0.36}    & \bf92.38\tinytiny{ 0.32}    & 91.79\tinytiny{ 0.34}
\\
\it test on \HUMAN     & 74.69\tinytiny{ 1.13}    & 73.67\tinytiny{ 1.24}    & 71.22\tinytiny{ 1.20}    & \bf75.56\tinytiny{ 1.23}    & 74.33\tinytiny{ 1.26}    & 71.64\tinytiny{ 1.23}
\\

\bottomrule
\end{tabular}
\\
\raggedright
\footnotesize
Each value is an aggreg. of 125 exp. (5 splits $\times$ 5 SimCLR seeds $\times$ 5 fine-tune seeds); 
(\textcolor{red}{*}) pair of augmentations used in \cite{horowicz2022imc-fewshotcl}.
\end{table}

In Table~\ref{tab:dropout-projection-layer}, we also annotate
the configuration that (we believe) was used in the \TARGETPAPER.
Specifically, the study reported results (only as figures)
characterising performance improvement when increasing the number
of samples for fine-tune training, and concluded that the best
performance was achieved when using 10 training samples, i.e.,
the scenario we selected for our evaluation. Yet, while
for \SCRIPT \TARGETAUTHORS wrote:
\begin{quotethepaper}
\footnotesize\
Our method
achieves 93.4\% accuracy with only 3 samples, and 94.5\% with
10 samples
\end{quotethepaper}
\noindent no specific values are reported for \HUMAN.  However,
Figure 4 of the paper clearly shows an accuracy of about $80\%$.

While performance are basically on par for \SCRIPT,
our results for \HUMAN\ are significantly lower than the previous evaluation.
We additionally observe that, for  contrastive learning with fine-tuning,
a  drop of performance from \SCRIPT to \HUMAN  is also reported in
the \TARGETPAPER---unlike for supervised training as  discussed earlier.
While the training methodologies are fundamentally different,
the underlying dataset and testing methodology are the same
(training with the pretraining partition and testing
on \SCRIPT and \HUMAN). Thus, the consistency between our
ML, supervised and contrastive learning campaigns is to
be expected, but we cannot comment on why \TARGETAUTHORS
observed the \SCRIPT-vs-\HUMAN gap only for the contrastive learning
experiments.\footnote{Authors did not provide more comments to us
about this aspect.}

\mypar{Takeaway}{\it
On the one hand, results are consistent and quantitatively aligned for \SCRIPT,
which confirms the interest for few shot contrastive learning and data augmentation.
On the other hand, results for \HUMAN\ are only qualitatively in agreement,
which calls for agreeing on a community-wide standard benchmark including multiple datasets.
}

\subsubsection{Extra results}

We conclude our analysis by reporting two complementary analysis
with respect to the \TARGETPAPER. First, we investigated
to which extent alternative pairs of augmentations
affect the fine-tuning performance. Namely, we considered
\emph{Time shift} and \emph{Change RTT} next to
\emph{Rotate} and \emph{Color jitter}, selected because
they achieved good positions in our ranking analysis.
Then we formed groups by either pairing time series with image transformations
or pairing the image transformations.
Results collected in Table~\ref{tab:augmentation-pairs} show that,
despite the punctual differences between pairs,
our observation on Table~\ref{tab:augmentation-at-loading} and the ranking analysis (Sec~\ref{sec:G1.2})
still holds---all pairs are \emph{qualitatively} equivalent.

\begin{table}%
\centering
\footnotesize
\caption{
Accuracy on 32$\times$32 flowpic when enlarging training set (without dropout).\label{tab:supervised_and_CT_entrire_trainset}}
\begin{tabular}
{
    @{} 
    r@{$\;\;$} 
    r@{$\;\;$}
    r@{$\;\;$}
    r@{$\;\;$}
    r
    @{}
}
\toprule 
  & & \multicolumn{1}{c}{\SCRIPT} & \multicolumn{1}{c}{\HUMAN}  & 
\\
\cmidrule(r){1-2} \cmidrule(r){3-4} 
\parbox[t]{2mm}{\multirow{7}{*}{\rotatebox[origin=c]{90}{Supervised}} }
& No augmentation         & 98.37\tinytiny{ 0.19}    & 72.95\tinytiny{ 0.96}
\\
& Rotate                  & 98.47\tinytiny{ 0.25}    & 73.73\tinytiny{ 1.09}
\\
& Horizontal flip         & 98.20\tinytiny{ 0.15}    & 74.58\tinytiny{ 1.16}
\\
& Color jitter            & 98.63\tinytiny{ 0.21}    & 72.47\tinytiny{ 1.02}
\\
& Packet loss             & 98.63\tinytiny{ 0.19}    & 73.43\tinytiny{ 1.25}
\\
& Time shift              & 98.60\tinytiny{ 0.22}    & 73.25\tinytiny{ 1.17}
\\
& Change RTT              & 98.33\tinytiny{ 0.16}    & 72.47\tinytiny{ 1.04}
\\
\midrule
& SimCLR + fine-tuning    & 93.90\tinytiny{ 0.74}    & 80.45\tinytiny{ 2.37}
\\
\bottomrule
\end{tabular}
\\
Each value is an aggregation of 20 experiments (20 different seeds)
\end{table}

Second, we expand the methodology used so far by quantifying
the effect of using a (pre)training set larger than 100 samples.
Specifically, we created 5 random 80/20 train/validation split using the full
pretraining partition, i.e., the dataset result imbalanced with
up to 1,532 training samples for the largest class and 473 for the smallest.
Table~\ref{tab:supervised_and_CT_entrire_trainset} reports
the results of the modeling campaign in both a supervised
and contrastive learning settings.
As expected, compared to Table \ref{tab:augmentation-at-loading} and Table \ref{tab:dropout-projection-layer},
enlarging the dataset is effective in improving performance in both settings.
In particular, for contrastive learning the gain is smaller for
\SCRIPT (+1.72\% on average) than for \HUMAN
(+5.76\% on average)---the latent space
created via contrastive learning is better
at mitigating the data shift.

\mypar{Takeaway}{\it
The transformations selected in the \TARGETPAPER  constitute
a good enough choice, although image transformations
cannot be fully ruled out based on our assessment.
This confirms that identifying the most suitable transformations
is tied to the input representation and datasets used, which remains an open problem.
Moreover, while  a very limited number of samples
can be enough for training models, the same scenarios can
benefit from more data---the selected augmentations alone are not
a final replacement for real input samples.
}

\begin{table}%
\centering
\footnotesize
\caption{
(G3) Data augmentation in supervised  setting on other  datasets. The top two transformation strategies for each datasets are in bold for visual purposes (not to imply statistically relevant conclusions).\label{tab:Supervised_other_dataset}}
\begin{tabular}
{
    @{} 
    r@{$\;\;$} 
    r@{$\;\;$}
    r@{$\;\;$}
    r@{$\;\;$}
    r@{$\;\;$} 
    r
    @{}
}
\toprule 
& 
\MIRAGEB  &  
\MIRAGEB &
\UTMOBILENET  & 
\MIRAGEA &
\\
Augmentation  &
 ($\geq$10pkts)  &
 ($\geq$1000pkts) &
 ($>$10pkts) &
 ($>$10pkts) &
\\
\cmidrule(r){1-1} \cmidrule(r){2-5}
No augmentation          & 90.97 \tinytiny{ 1.15}     & 83.35 \tinytiny{ 3.13}     & 79.82 \tinytiny{ 1.53}     & 69.91 \tinytiny{ 1.57}
\\
Rotate                   & 88.25 \tinytiny{ 1.20}     & \bf87.32 \tinytiny{ 2.24}     & 79.45 \tinytiny{ 1.28}     & 60.35 \tinytiny{ 1.17}
\\
Horizontal flip          & 91.90 \tinytiny{ 0.84}     & 83.82 \tinytiny{ 2.26}     & 80.03 \tinytiny{ 1.33}     & 69.78 \tinytiny{ 1.28}
\\
Color jitter             & 89.77 \tinytiny{ 1.16}     & 81.40 \tinytiny{ 3.62}     & 78.68 \tinytiny{ 2.14}     & 67.00 \tinytiny{ 1.11}
\\
Packet loss              & 92.34 \tinytiny{ 1.10}     & 87.19 \tinytiny{ 2.52}     & 72.07 \tinytiny{ 1.73}     & 67.55 \tinytiny{ 1.46}
\\
Time shift               & \bf92.80 \tinytiny{ 1.21}     & 86.73 \tinytiny{ 3.88}     & \bf81.91 \tinytiny{ 2.12}     & \bf70.33 \tinytiny{ 1.26}
\\
Change RTT               & \bf93.75 \tinytiny{ 0.83}     & \bf91.48 \tinytiny{ 2.12}     & \bf81.32 \tinytiny{ 1.54}     & \bf74.28 \tinytiny{ 1.22}
\\
\bottomrule
\end{tabular}
\\
Each value is aggregation of 15 experiments (5 splits $\times$ 3 seeds).
\end{table}

\subsection{Replicating data augmentation on other datasets (G3)}
\subsubsection{Approach}
Given that we observed only small performance differences among the augmentations,
we extended the \TARGETPAPER by replicating the analysis 
using other three datasets, namely \MIRAGEA, \MIRAGEB and \UTMOBILENET.
Based on the results displayed in Table~\ref{tab:supervised_and_CT_entrire_trainset}, we opted for
a traditional 80/10/10 train/validation/test using all samples available for each class,
i.e., we removed the constraint of using 100 samples per class as in Table~\ref{tab:augmentation-at-loading}.
This is a compromise dictated by the differences among the datasets.
In particular, as shown in Table~\ref{tab:datasets}, the filtering significantly reduces the number
of samples per class, especially for the smallest class.
Hence, rather than removing the very small classes, we preferred to use a split
preserving the original imbalance of the data.
We argue that this is reasonable considering that the
question we were targeting was about
the \emph{importance of the augmentation functions} which is per-se
to be decoupled from datasets samples count.
Moreover, we restricted our analysis to the supervised scenario only.
It follows that our analysis can be considered as an \emph{upper bound}
of what can be achieved when considering less training data and/or
via contrastive learning.
Since the training and testing datasets are imbalanced in this scenario,
we measure performance via an F1 score (rather than using accuracy
as done before).

\subsubsection{Results}
For each dataset we used the architectures and settings as in Sec.~\ref{sec:flowpic:dataaugment}.
Table~\ref{tab:Supervised_other_dataset} and Figures
\ref{fig:CD_replicability_together}-\ref{fig:CD_replicability_isolated}
collect our results.
Extending the analysis to more datasets allows us to better appreciate differences between the impact of each augmentation.
First of all, while the maximum gap between augmentations in Table~\ref{tab:augmentation-at-loading} is (on average) 3.22\%,
this is now 13.93\% (occurring for \MIRAGEA).
Despite the larger differences, the analysis confirms Change RTT and Time shift as
the best performing augmentations across all datasets.
Differently from the previous analysis, Fig.~\ref{fig:CD_replicability_together}-\ref{fig:CD_replicability_isolated}
highlight how the two functions are significantly better than the others,
yet still not statistically different from each other.

\mypar{Takeaway}{\it
Our results confirm the benefit of data augmentations and
validate the selection of Change RTT and Time shift as in the \TARGETPAPER.
}

\begin{figure}[t]
    \centering
    \includegraphics[width=\columnwidth]{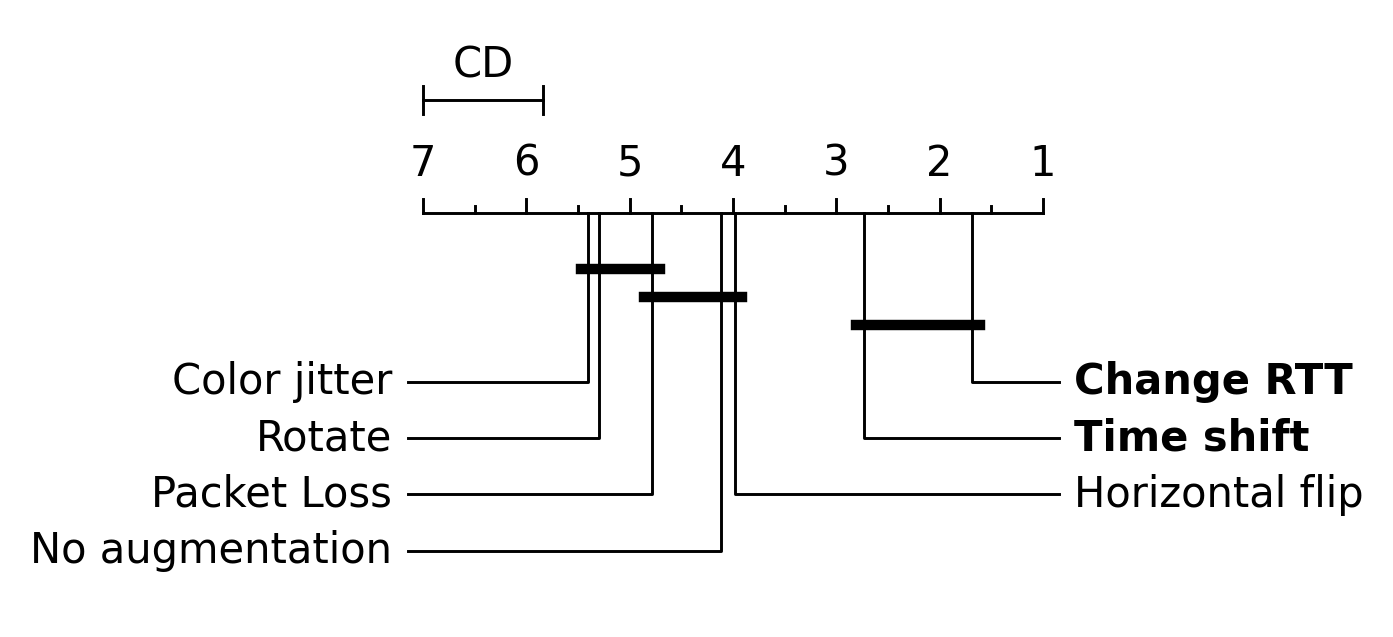}
    \caption{Critical distance plot of the accuracy obtained with each augmentation across the four tested datasets.} 
    \label{fig:CD_replicability_together}
\end{figure}

\section{Conclusions\label{sec:conclusions}}

In this paper we reproduced and replicated the methodology of \cite{horowicz2022imc-fewshotcl} which investigated noteworthy DL
methodologies (few-shot learning, self-supervision
via contrastive learning and data augmentation) on TC.
These methods are particularly appealing as they allow for learning from a few samples and transferring models across datasets.

Summarizing our analysis, we have been able to \emph{qualitatively} reproduce most of the original results,
so we confirm the interest in few-shot contrastive learning and data augmentation.
At the same time, our modeling campaigns found unexpected \emph{quantitative}
discrepancies that we rooted in data shifts in the \UCDAVIS dataset (undetected in the \TARGETPAPER).

Another remarkable consideration can be gathered by contrasting our reproducibility vs replicability results.
Indeed, the reproducibility results on \UCDAVIS show little statistical significance
in the differences among the proposed data augmentation techniques---just by reproducing the study
on \UCDAVIS alone would therefore have not allowed us to validate \TARGETAUTHORS's choices.
Conversely, by replicating the methodology on three additional datasets, we gathered evidence that finally validated Change RTT and Time Shift as more beneficial than other augmentations for the flowpic input representation.

We also acknowledge some limitations in our replication. For instance, while we studied augmentations in a supervised setting, we leave as future work their assessment in a contrastive learning setting paired with few shot fine-tuning.
Indeed, such a study should consider the variety of contrastive learning approaches including \emph{supervised} contrastive learning methods such as SupCon~\cite{khosla2020neurips-supcon}.

\begin{figure}[t]
    \centering
    \includegraphics[width = \columnwidth]{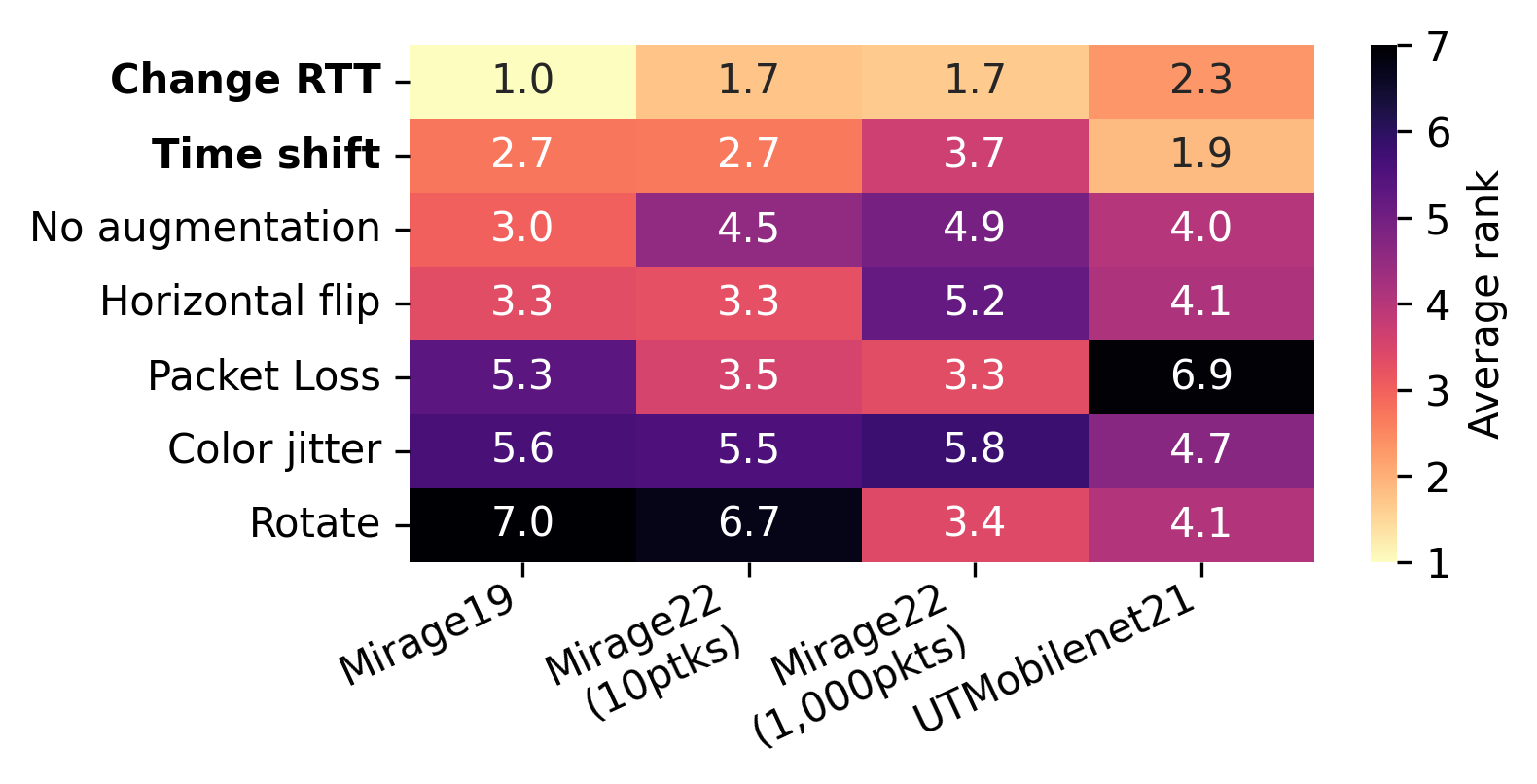}
    \caption{Average rank obtained per augmentation and dataset. Ranks closer to 1 indicate a better performance.}
    \label{fig:CD_replicability_isolated}
\end{figure}

Lastly, in the context of network ML studies, we underline the need to agree on a broader set of benchmarks as other communities (e.g., CV and NLP) are doing more systematically, which can only improve the quality of the gathered knowledge.
To support this future direction, we make available multiple artifacts in the form of code
(besides our modeling framework, we contribute scripts related to the modeling campaign and all post-processing to generate reports and figures inhere contained) and data (both trained models and related logs, as well as the dataset splits used for training and testing). As described in App.~\ref{app:artifacts}, artifacts are also complemented by a website providing documentation (e.g., guides on how to run the experiments, stats about the datasets).
We believe that TC is in need of a reference framework binding datasets with modeling tools.
We hope the research community can take advantage of our work and/or be inspired toward improving current practices.

\bibliographystyle{ACM-Reference-Format}
\bibliography{reference}

\begin{figure*}
    \centering
    \includegraphics[width=\textwidth]{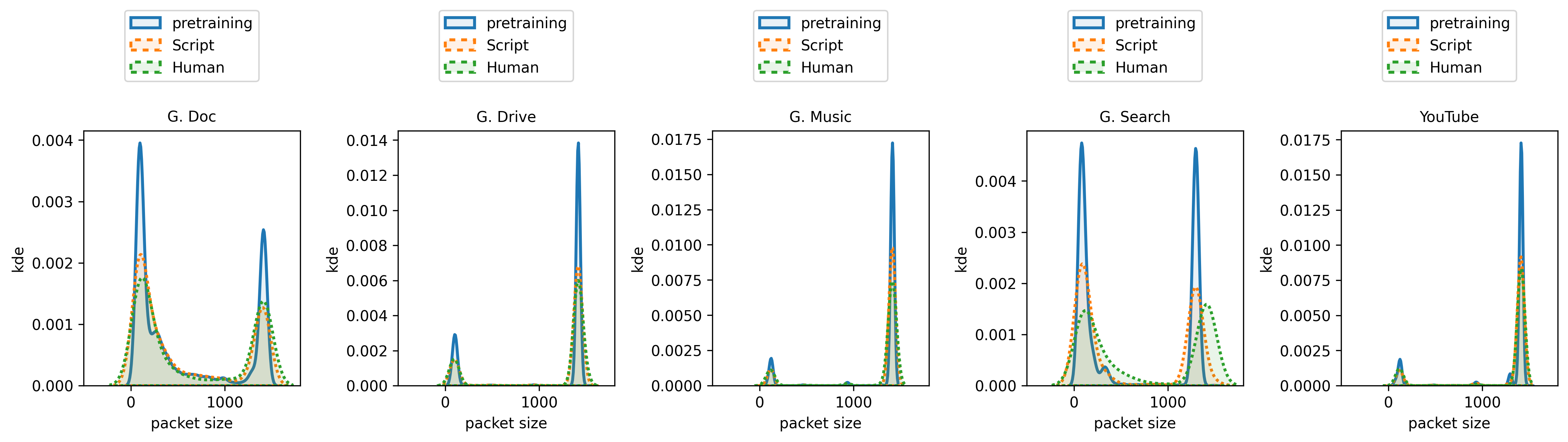}
    \caption{Kernel density estimation of the per-class packet size distributions. Notice the distribution shift for Google search.}
    \label{fig:kde}
\end{figure*}

\appendix

\section{Ethics}
This work makes use of only publicly available data.
Although experiments might have included end users,
no individuals were monitored in the measurement campaigns.
Thus, no ethical concerns are associated with this study.

\section{Artifacts}
\label{app:artifacts}
All the material created for this paper is made available to
the research community. This includes our modeling framework,
namely \texttt{tcbench},
all data created in our modeling campaigns (the models themselves
and all related logs) and the curated datasets described in Sec.~\ref{sec:datasets}.
For more details on the artifacts, how to run our modeling campaigns,
notebooks for recreating tables and figures in this paper,
data curation, etc.,
please visit \url{https://tcbenchstack.github.io/tcbench/}.

\section{Layout of DL Network Architectures}\label{app:architectures}

We list here our implementation of the
network architectures for the
32$\times$32 flowpic resolution (see also
Fig.7 in \cite{horowicz2022imc-fewshotcl}).
The Listings~\ref{net:supervised-withdropout}-\ref{net:finetune} at the
end of this appendix are obtained via
the \texttt{torchsummary} python package.
For code flexibility, our
architectures are designed to use
Pytorch \texttt{nn.Identity()} modules
to mask out layers that are not needed
from a given architecture.
When this masking is applied, our
training framework takes care of
recreating the network optimizers to
reflect the architecture modifications.

\section{Investigating root cause of G1 discrepancies}\label{quote: icdm-datashit-explain}
We were clearly surprised by the $\approx$20\%
classification accuracy performance gap for the \HUMAN test split.
In fact, in the \TARGETPAPER the two testing partitions
have almost on par performance.
As mentioned before, when reaching out to \TARGETAUTHORS
we received delayed and short/partial answers so
we based our analysis mostly on the content of the paper.
By browsing the web, we also found a git repository~\cite{unofficialgit}
that later on was confirmed to have been created by the first author of the \TARGETPAPER.
Unfortunately, this repository only contains code
related to the contrastive learning part of the paper
(i.e., it only pre-trains a model to later investigate
its latent space using a t-SNE projection in two dimensions).
Moreover, the network architecture used significantly differs from the one described in
the \TARGETPAPER
(e.g., different activation functions, no dropout is used) and also adopts a
cosine annealing learning rate scheduler (not mentioned in the original
publication). Lastly, the data loading policies are blending flows
between the three partitions of the \UCDAVIS dataset (hence breaching the
evaluation protocol defined in the paper) and we found also
some of the flows in the test set to be included in the training set.
In a nutshell, the repository was of no use to address our questions.

We therefore performed more analysis of the \UCDAVIS
paying particular attention to details reported in \cite{rezaei2019ICDM-ucdavis}
which introduces the \UCDAVIS dataset.

\subsection{Our analysis of the \UCDAVIS dataset}\label{app:gap-dataset}
Next to Fig~\ref{fig:ucdavis-flowpic}, Fig.~\ref{fig:kde}
provides a more compelling argument about the presence of the
data shift by showing the Kernel Density Estimation (KDE)
of the per-class
packet size distribution across all samples
in the three partitions of \UCDAVIS.
While \SCRIPT is perfectly overlapped with the
pretraining split, \emph{Google search} for \HUMAN
has an evident shift, which indeed matches
the previous observations in Fig.~\ref{fig:ucdavis-flowpic}.

\subsection{\UCDAVIS dataset analysis from \cite{rezaei2019ICDM-ucdavis}}\label{app:gap-ucdavis}

We next looked at other sources of information which
could help us to
exclude problems from our analysis.  To the best of our knowledge, the only study that
investigates both \SCRIPT and \HUMAN splits
with a reference per-class breakdown is \cite{rezaei2019ICDM-ucdavis},
i.e., the same study that introduces the \UCDAVIS dataset.
The authors wrote:
\begin{quotethepaper}
\footnotesize\it
To study whether automatically generated data with script represents human interaction, we capture 15 flows for each class from interactions of real
humans in those 5 Google services. We only use this dataset to test the same
model described above. Fig. 3(b) illustrates the performance metrics. Interestingly, accuracy of the Google search and Google document have not changed
significantly. However, the accuracy of Google drive, Youtube, and Google music drop up to 7\%. This depends on how much human interactions can change
the traffic pattern, which is class-dependent. Moreover, there are some actions,
such as renaming a file or moving files in Google drive, that our scripts do not
perform. So, these patterns are not available during re-training. This shows the
limitations of datasets and studies [14, 8, 3] that only use scripts to capture data.
\end{quotethepaper}
Interestingly, they reported no problem with \emph{Google doc} and \emph{Google search}, yet they acknowledged the presence of data-shift due to the way the
dataset was collected.
However this information alone cannot help us explaining the
discrepancies we observe with respect to the \TARGETPAPER.
In fact, \cite{rezaei2019ICDM-ucdavis} relies
on packet time series augmented via sampling;
in contrast, a flowpic is a ``summary'' which
aggregates patterns over time and does
not consider traffic direction---the two studies have
intrinsically different input.

Also~\cite{towhid-netsoft22-boyl} uses the \UCDAVIS dataset
but authors combine all partitions together, i.e., they
do not follow the training/testing protocol of \cite{rezaei2019ICDM-ucdavis,horowicz2022imc-fewshotcl}.

\begin{table}%
\centering
\footnotesize
\caption{
Macro-average accuracy with different re-training dataset and different sampling methods.\label{tab:icdm19_fig2a}}
\begin{tabular}{
    @{}
    r@{$\;\;$} 
    r@{$\;\;$}
    r@{$\;\;$}
    r@{$\;\;$}
    r@{$\;\;$}
    r@{$\;\;$} 
    r@{$\;\;$}
    r
    @{}
}
\toprule 
& \multicolumn{3}{c}{\bf  from \cite{rezaei2019ICDM-ucdavis} Fig. \ref{fig:icdm19_fig2a_paper}} & \multicolumn{3}{c}{\bf  ours}  
\\
\cmidrule(r){2-4} \cmidrule{5-7}
\bf finetune &
\multicolumn{3}{c}{\bf Sampling$^{\,\dagger}$} &
\multicolumn{3}{c}{\bf Sampling} &
\\
\bf on & 
\bf Fixed & \bf Rand & \bf Incre &
\bf Fixed & \bf Rand & \bf Incre &
\\
\cmidrule(r){2-4} \cmidrule{5-7}
\SCRIPT   &  
\textcolor{green}{92.28  (b)} & \textcolor{red}{92.28  (c)} &  \textcolor{blue}{95.59  (a)}  &
87.11  \tinytiny{0.09}  & 94.63 \tinytiny{0.02}  & 96.22 \tinytiny{0.01} & \\
\HUMAN    & 
-  & -  & -  & 
 82.60 \tinytiny{0.03}  &  87.29\tinytiny{0.04}  &  92.56 \tinytiny{0.03} & \\
\bottomrule
\end{tabular}
\tiny
$\dagger$ ``Fixed'': Fixed step sampling; ``Rand'': Random sampling; ``Incre'': Incremental sampling.
\end{table}

\subsection{Reproduction of \cite{rezaei2019ICDM-ucdavis} on \UCDAVIS}\label{app:gap-reproduce}
Finally, to rule out errors in our execution,
we leveraged \cite{rezaei2019ICDM-ucdavis}
code artifacts.\footnote{\url{https://github.com/shrezaei/Semi-supervised-Learning-QUIC-}}
To ensure that the \UCDAVIS dataset we used was intact and correct and to analyze the impact of data shift between \SCRIPT and \HUMAN partitions, 
we reproduced some the results of \cite{rezaei2019ICDM-ucdavis} using the available repository,
yet using our curated version of the \UCDAVIS dataset (which
we reiterate was just reworking the original
CSV files into a monolithic parquet format).
In \cite{rezaei2019ICDM-ucdavis}, for each flow,
3 different sampling methods (i.e., random sampling, fixed step sampling, and incremental sampling) are applied respectively up to 100 times to generate multiple short ``subflow'' time-series, thus augmenting the data set.
For self-supervised pre-training on the entire pre-training partition, the authors used a statistical features regression task.
For supervised fine-tuning, 3 linear layers are stacked as classifier for the classification task and they are trained with up to 20 labeled flows.
While the fine-grained details of the training differ compared to our study and the \TARGETPAPER,
at a high level these three studies share the same aim, i.e.,
the first pre-train (on the pretraining partition) and then fine-tune (on the two test partitions).

\begin{figure}[t]%
    \centering
    \includegraphics[width=0.3\textwidth]{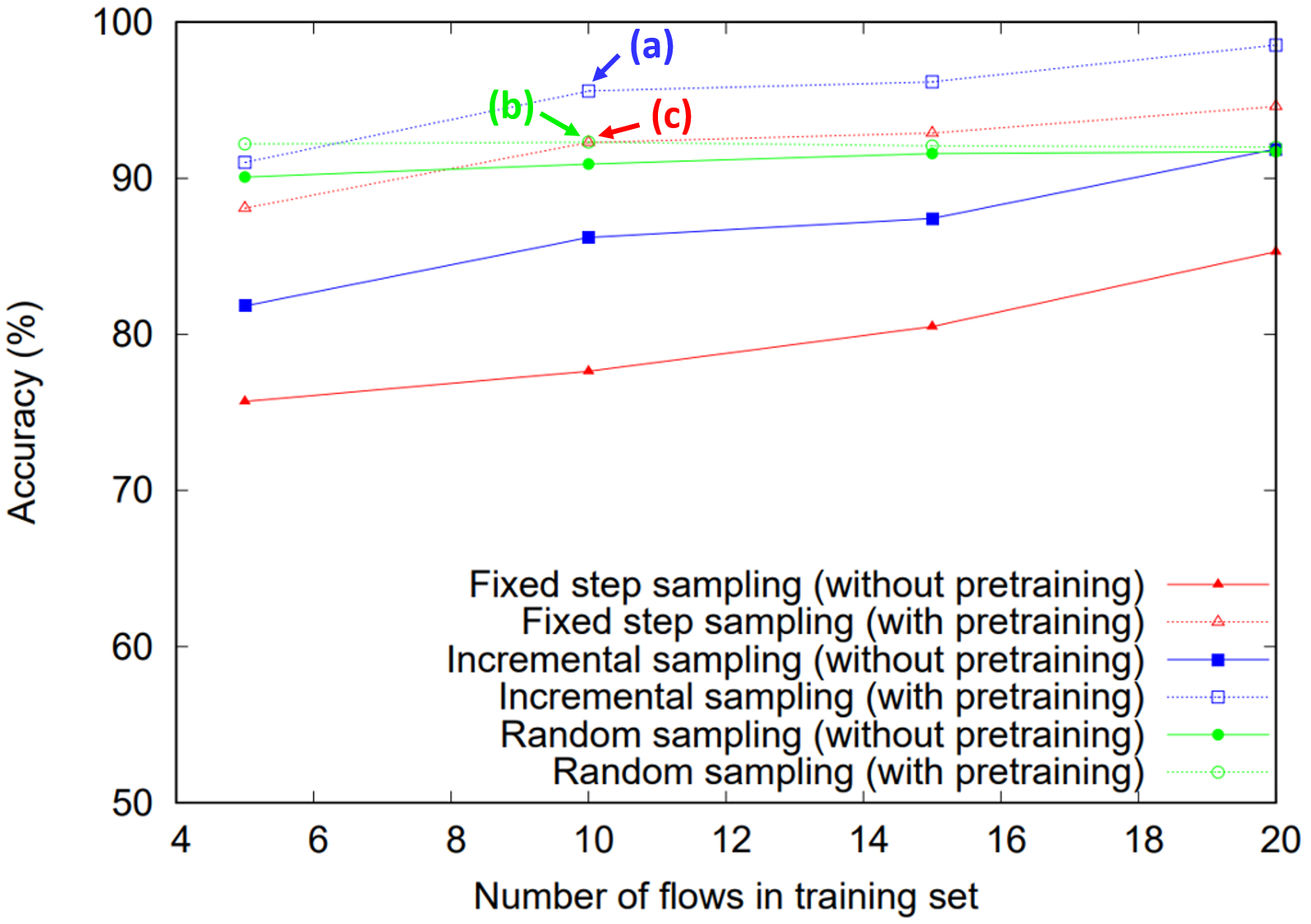}
    \caption{Accuracy on \SCRIPT with different sampling methods\cite{rezaei2019ICDM-ucdavis}.}
    \label{fig:icdm19_fig2a_paper}
\end{figure}

\begin{figure}[t]
    \centering
    \subfloat[\centering from \cite{rezaei2019ICDM-ucdavis}]{{\includegraphics[width=0.24\textwidth]{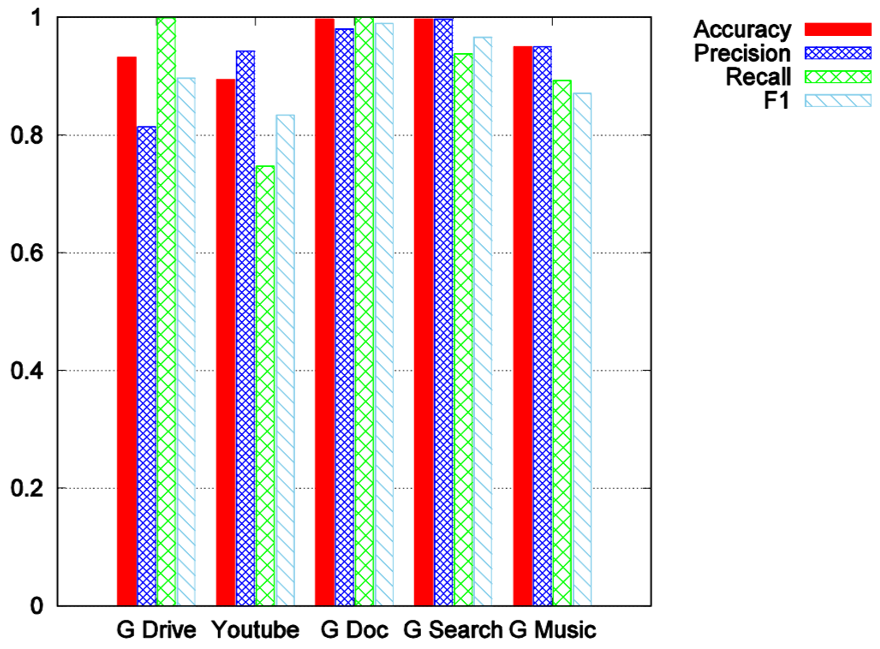} }}%
    \subfloat[\centering ours]{{\includegraphics[width=0.255\textwidth]{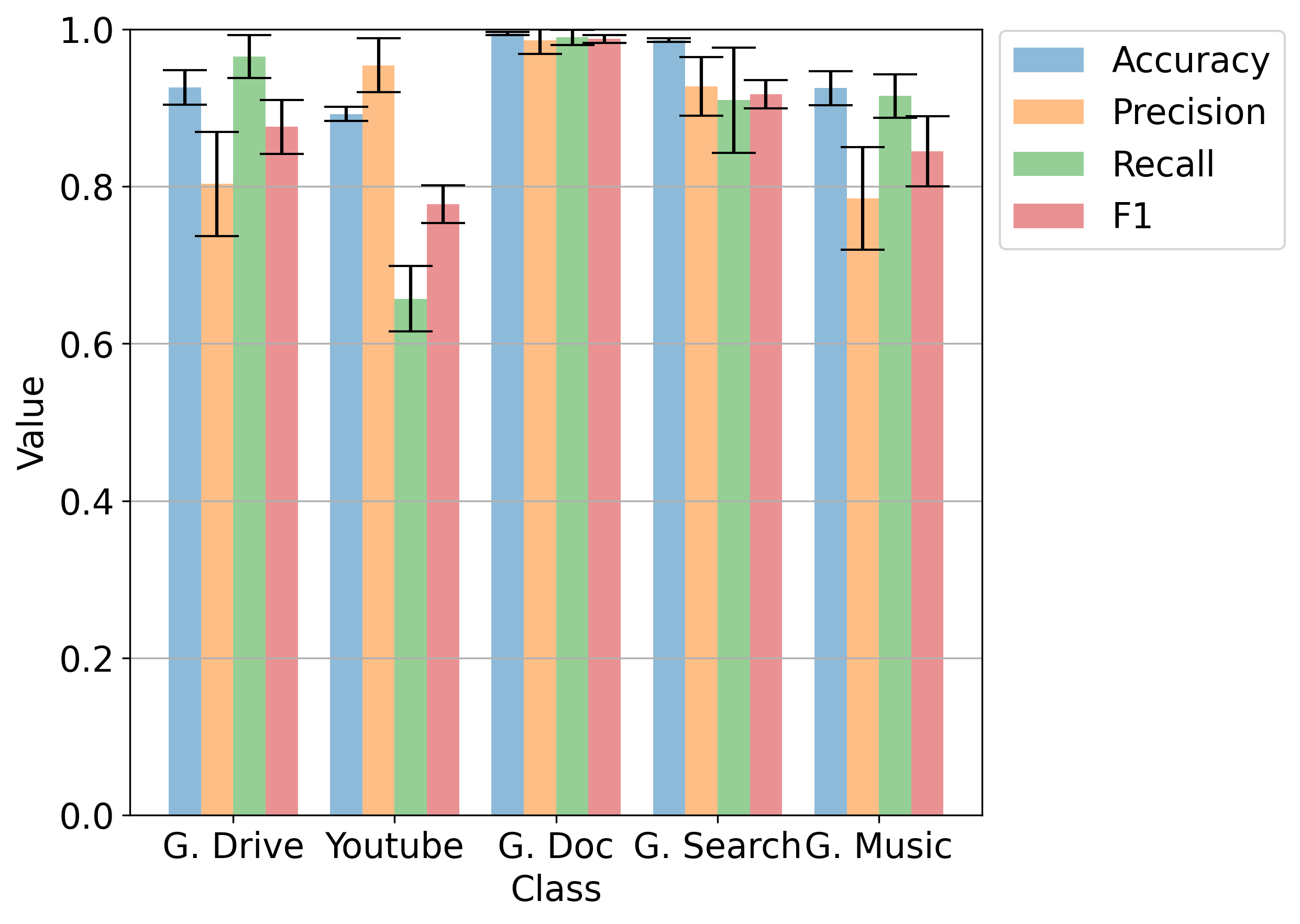} }}%
    \caption{\centering Replicating per-class accuracy on \HUMAN.}%
    \label{fig:icdm19_fig3b}%
\end{figure}

Table \ref{tab:icdm19_fig2a} reports the performance when fine-tuning with 10 samples.
In \cite{rezaei2019ICDM-ucdavis} the performance is only measured on \SCRIPT and is only reported as a figure without numeric annotation (see (a), (b), (c) in Fig.~\ref{fig:icdm19_fig2a_paper}, based on which
we inferred the values reported on the left side of Table \ref{tab:icdm19_fig2a}).
On the right side we reported the results from our modeling campaign
using the reference git repository.
Overall the accuracy on \SCRIPT has differences in the range 0.68-5.17\%
(much smaller than the $\approx$20\% accuracy gap under investigation), and we can also confirm their results, i.e.,
\emph{incremental sampling is the best strategy} for the method reported in \cite{rezaei2019ICDM-ucdavis}.
For \HUMAN instead we detect a $3.66-7.34\%$ drop with respect to \SCRIPT.
A similar drop is reported in \cite{rezaei2019ICDM-ucdavis} for incremental sampling
when fine-tuned on 20 \SCRIPT flows and tested on \HUMAN
(i.e., in a transfer learning setting) which we were also
able to replicate in Fig. \ref{fig:icdm19_fig3b}. 
Their reasoning for such differences is quoted in App.~\ref{app:gap-ucdavis}.
Overall, this evaluation is in line with the results of~\cite{rezaei2019ICDM-ucdavis}
and shows that our preprocessing of \UCDAVIS is not responsible
for the data shift we observed.

\section{Impact of dropout in a supervised setting\label{app:dropout-in-supervised}}

To assess the impact of dropout in a supervised setting, we performed an ablation study
using different test sets, resolutions, and augmentations for 32$\times$32 and 1500$\times$1500 resolutions
with the same campaign settings described in Sec.~\ref{sec:flowpic:dataaugment} (i.e., 15 experiments in
each configuration).
Fig.~\ref{fig:supervised_dropout_std} shows the results as
boxplots (with whiskers at the 95-th percentile)
of the difference between the accuracy when using
dropout with respect to when not using dropout.
In other words, dropout would be justified if the boxplots
would fall on the positive size of the y-axis.
Conversely, across all scenarios, the boxplots
are centered around zero with no evident patterns
across augmentations. Overall, we concluded that
of dropout does not play a role and its adoption
(as required by the \TARGETPAPER) is weakly motivated.

\begin{figure}[t]
    \centering
    \includegraphics[width=\columnwidth]{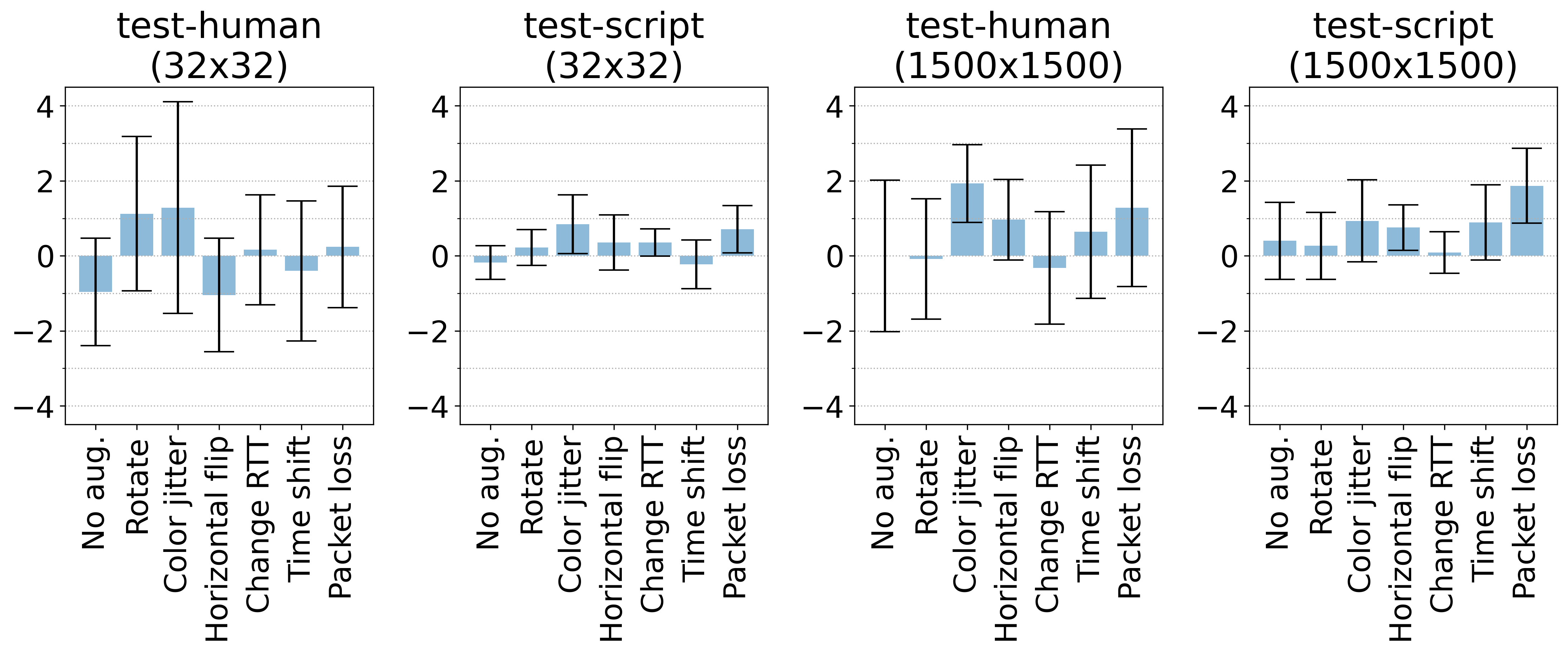}
    \caption{Boxplots of the accuracy difference between models with dropout and without dropout
    in supervised learning across different augmentations.}
    \label{fig:supervised_dropout_std}
\end{figure}

\section{Comparison of augmentations performance across flowpic sizes}\label{appendix:tukey}

In order to perform the analysis found in section \ref{sec:G1.2}, three sets of experiments were available, corresponding to the different flowpic resolutions used: 32$\times$32, 64$\times$64 and 1500$\times$1500. If possible, it would be desirable to group the three sets into a single analysis, as that increases the $N$ in the Critical Distance calculation, which reduces the CD's width and allows us to better differentiate between augmentations. However, first we had to ensure that the augmentations performance across sets are similar. To do so, we treated each flowpic resolution as a classifier and compared their paired performance distributions using a posthoc Tukey test, which calculates whether each resolution's performance can be assumed to be significantly different from each other or not. This test's results are shown on Table \ref{tab:flowpic_tukey} with the p-values for each comparison. We used a significance level of 0.05, i.e., we can assume significant differences between resolutions if their p-value is smaller than 0.05. There are two populations for which augmentations perform in a similar way: 32$\times$32 and 64$\times$64, with 1500$\times$1500 being clearly different from the other two.
Based on these results, we joined the 32$\times$32 and 64$\times$64 populations for our analysis in section \ref{sec:G1.2}.

\begin{table}[h]
\centering
\footnotesize
\caption{
Performance comparison across augmentations for different flowpic sizes. P-values extracted from Tukey's post-hoc test at a 0.05 significance level.}
\begin{tabular}{cccc}
\hline
Flowpic resolution & Flowpic resolution & p-value & Is Different? \\ \hline
32$\times$32                & 64$\times$64                & 0.57               & No         \\
32$\times$32                & 1500$\times$1500            & \num{1.93e-6}      & Yes        \\
64$\times$64                & 1500$\times$1500            & \num{1.04e-8}      & Yes        \\ \hline
\end{tabular}\label{tab:flowpic_tukey}
\end{table}

\newpage

\begin{lstlisting}[caption=Supervised network (with dropout)., label={net:supervised-withdropout}]
flowpic_dim: 32
num_classes: 5
with_dropout: True
----------------------------------------------
  Layer (type)        Output Shape    Param #
==============================================
      Conv2d-1     [-1, 6, 28, 28]        156
        ReLU-2     [-1, 6, 28, 28]          0
   MaxPool2d-3     [-1, 6, 14, 14]          0
      Conv2d-4    [-1, 16, 10, 10]      2,416
        ReLU-5    [-1, 16, 10, 10]          0
   Dropout2d-6    [-1, 16, 10, 10]          0
   MaxPool2d-7      [-1, 16, 5, 5]          0
     Flatten-8           [-1, 400]          0
      Linear-9           [-1, 120]     48,120
       ReLU-10           [-1, 120]          0
     Linear-11            [-1, 84]     10,164
       ReLU-12            [-1, 84]          0
  Dropout1d-13            [-1, 84]          0
     Linear-14             [-1, 5]        425
==============================================
Total params: 61,281
Trainable params: 61,281
Non-trainable params: 0
----------------------------------------------
Input size (MB): 0.00
Forward/backward pass size (MB): 0.13
Params size (MB): 0.23
Estimated Total Size (MB): 0.36
\end{lstlisting}

\begin{lstlisting}[caption=Supervised network (without dropout)., label={net:supervised-nodropout}]
flowpic_dim: 32
num_classes: 5
with_dropout: False
---------------------------------------------
  Layer (type)        Output Shape   Param #
=============================================
      Conv2d-1     [-1, 6, 28, 28]       156
        ReLU-2     [-1, 6, 28, 28]         0
   MaxPool2d-3     [-1, 6, 14, 14]         0
      Conv2d-4    [-1, 16, 10, 10]     2,416
        ReLU-5    [-1, 16, 10, 10]         0
    Identity-6    [-1, 16, 10, 10]         0 <-- masked
   MaxPool2d-7      [-1, 16, 5, 5]         0
     Flatten-8           [-1, 400]         0
      Linear-9           [-1, 120]    48,120
       ReLU-10           [-1, 120]         0
     Linear-11            [-1, 84]    10,164
       ReLU-12            [-1, 84]         0
   Identity-13            [-1, 84]         0 <-- masked
     Linear-14             [-1, 5]       425
=============================================
Total params: 61,281
Trainable params: 61,281
Non-trainable params: 0
---------------------------------------------
Input size (MB): 0.00
Forward/backward pass size (MB): 0.13
Params size (MB): 0.23
Estimated Total Size (MB): 0.36
\end{lstlisting}

\begin{lstlisting}[caption=SimCLR pre-train (small projection layer)., label={net:simclr}]
flowpic_dim: 32
num_classes: 5, 
projection_layer_dim: 30
with_dropout: False
---------------------------------------------
  Layer (type)        Output Shape   Param #
=============================================
      Conv2d-1     [-1, 6, 28, 28]       156
        ReLU-2     [-1, 6, 28, 28]         0
   MaxPool2d-3     [-1, 6, 14, 14]         0
      Conv2d-4    [-1, 16, 10, 10]     2,416
        ReLU-5    [-1, 16, 10, 10]         0
    Identity-6    [-1, 16, 10, 10]         0
   MaxPool2d-7      [-1, 16, 5, 5]         0
     Flatten-8           [-1, 400]         0
      Linear-9           [-1, 120]    48,120
       ReLU-10           [-1, 120]         0
     Linear-11           [-1, 120]    14,520 <- proj layer 1
       ReLU-12           [-1, 120]         0
   Identity-13           [-1, 120]         0
     Linear-14            [-1, 30]     3,630 <- smaller proj layer
=============================================
Total params: 68,842
Trainable params: 68,842
Non-trainable params: 0
---------------------------------------------
Input size (MB): 0.00
Forward/backward pass size (MB): 0.13
Params size (MB): 0.26
Estimated Total Size (MB): 0.39
\end{lstlisting}

\newpage

\begin{lstlisting}[caption=SimCLR pre-train (large projection layer)., label={net:simclr}]
flowpic_dim: 32
num_classes: 5, 
projection_layer_dim: 84
with_dropout: False
---------------------------------------------
  Layer (type)        Output Shape   Param #
=============================================
      Conv2d-1     [-1, 6, 28, 28]       156
        ReLU-2     [-1, 6, 28, 28]         0
   MaxPool2d-3     [-1, 6, 14, 14]         0
      Conv2d-4    [-1, 16, 10, 10]     2,416
        ReLU-5    [-1, 16, 10, 10]         0
    Identity-6    [-1, 16, 10, 10]         0
   MaxPool2d-7      [-1, 16, 5, 5]         0
     Flatten-8           [-1, 400]         0
      Linear-9           [-1, 120]    48,120
       ReLU-10           [-1, 120]         0
     Linear-11           [-1, 120]    14,520 <- proj layer 1
       ReLU-12           [-1, 120]         0
   Identity-13           [-1, 120]         0
     Linear-14            [-1, 84]    10,164 <- larger proj layer
=============================================
Total params: 75,376
Trainable params: 75,376
Non-trainable params: 0
---------------------------------------------
Input size (MB): 0.00
Forward/backward pass size (MB): 0.13
Params size (MB): 0.29
Estimated Total Size (MB): 0.42
\end{lstlisting}

\begin{lstlisting}[caption=Fine-tune network., label={net:finetune}]
flowpic_dim: 32
num_classes: 5, 
projection_layer_dim: 30
with_dropout: False
---------------------------------------------
  Layer (type)        Output Shape   Param #
=============================================
      Conv2d-1     [-1, 6, 28, 28]       156
        ReLU-2     [-1, 6, 28, 28]         0
   MaxPool2d-3     [-1, 6, 14, 14]         0
      Conv2d-4    [-1, 16, 10, 10]     2,416
        ReLU-5    [-1, 16, 10, 10]         0
    Identity-6    [-1, 16, 10, 10]         0
   MaxPool2d-7      [-1, 16, 5, 5]         0
     Flatten-8           [-1, 400]         0
      Linear-9           [-1, 120]    48,120
       ReLU-10           [-1, 120]         0
   Identity-11           [-1, 120]         0 <- masked
   Identity-12           [-1, 120]         0 <- masked
   Identity-13           [-1, 120]         0 <- masked
     Linear-14             [-1, 5]       605 <- final classifier
=============================================
Total params: 51,297
Trainable params: 51,297
Non-trainable params: 0
---------------------------------------------
Input size (MB): 0.00
Forward/backward pass size (MB): 0.13
Params size (MB): 0.20
Estimated Total Size (MB): 0.33
\end{lstlisting}

\raggedend

\end{document}